\documentclass[runningheads]{llncs}
\usepackage[T1]{fontenc}
\usepackage{graphicx}
\usepackage{booktabs}
\usepackage[misc]{ifsym}
\newcommand{\corr}{(\Letter)}

\usepackage[utf8]{inputenc} 
\usepackage{hyperref}       
\usepackage{amsfonts}       
\usepackage{nicefrac}       
\usepackage{microtype}      
\usepackage{xcolor}         

\usepackage{amsmath}
\usepackage{amssymb}
\usepackage{dsfont}
\usepackage{enumerate}
\usepackage{placeins}
\usepackage{algorithm}
\usepackage{algpseudocode}
\usepackage{subcaption}
\usepackage{multirow}

\newtheorem{method}{Method}
\newtheorem{assumption}{Assumption}

\DeclareMathOperator*{\argmax}{arg\,max}
\DeclareMathOperator*{\argmin}{arg\,min}

\newcommand*\bDelta{\boldsymbol{\Delta}}

\newcommand*\btheta{\boldsymbol{\theta}}

\newcommand*\caL{\mathcal{L}}
\newcommand*\bD{\mathbf{D}}

\newcommand*\bg{\mathbf{g}}
\newcommand*\bG{\mathbf{G}}

\newcommand*\bH{\mathbf{H}}
\newcommand*\bI{\mathbf{I}}

\newcommand*\bK{\mathbf{K}}

\newcommand*\bT{\mathbf{T}}
\newcommand*\bU{\mathbf{U}}
\newcommand*\bu{\mathbf{u}}
\newcommand*\bv{\mathbf{v}}

\newcommand*\bfeta{\boldsymbol{\eta}}

\newcommand*\metH{\bar{\mathbf{H}}}

\newcommand*\dd{\mathrm{d}}

\title{Gathering and Exploiting Higher-Order Information
	when Training Large Structured Models}
\titlerunning{Higher-Order Information for Large Structure Models}

\author{Pierre Wolinski\orcidID{0000-0003-1007-0144} \corr}
\authorrunning{P. Wolinski}
\institute{LAMSADE, 
	Paris-Dauphine University, 
	PSL University, 
	CNRS, 75016 Paris, 
	France \email{pierre.wolinski@dauphine.psl.eu}}

\begin{document} \sloppy 

\maketitle

\begin{abstract}
When training large models, such as neural networks, 
the full derivatives of order 2 and beyond are usually inaccessible,
due to their computational cost.
Therefore, among the second-order optimization methods, it is common
to bypass the computation of the Hessian by using 
first-order information, such as the gradient of the parameters (e.g., quasi-Newton methods)
or the activations (e.g., K-FAC).

In this paper, we focus on the exact and explicit computation
of projections of the Hessian and higher-order derivatives on
well-chosen subspaces relevant for optimization.
Namely, for a given partition of the set of parameters, 
we compute tensors that can be seen as
``higher-order derivatives according to the partition'',
at a reasonable cost as long as the number of subsets of 
the partition remains small.

Then, we give some examples of how these tensors can be used.
First, we show how to compute a learning rate per subset of parameters, which can
be used for hyperparameter tuning.
Second, we show how to use these tensors at order 2
to construct an optimization method that uses information contained 
in the Hessian.
Third, we show how to use these tensors at order 3 (information contained
in the third derivative of the loss)
to regularize this optimization method. The resulting training step has several interesting properties,
including: it takes into account long-range interactions
between the layers of the trained neural network, 
which is usually not the case in similar methods (e.g., K-FAC);
the trajectory of the optimization is invariant under 
affine layer-wise reparameterization.
\end{abstract}

\section{Introduction}

In machine learning, computing the derivatives of the loss at various
orders is challenging when using large models, such as neural networks.
While the first-order derivative is relatively cheap to compute and 
easy to use to train neural networks, things get difficult
when it comes to higher-order derivatives. 
In particular, Hessian-based training algorithms such as Newton's method
are very expensive to use on large models.
Therefore, the study of the Hessian
of a loss according to many parameters
has become a research area in its own right.
For derivatives of order 3 and higher,
the situation is even worse: their exact computation is far more expensive than the Hessian,
and only a few optimization algorithms use them.

\paragraph{Main contribution: extracting higher-order information.}
Formally, we study a loss $\caL$ to be minimized according to
a vector of parameters $\btheta \in \mathbb{R}^P$.
Thus, the order-$d$ derivative of $\caL$ at a given point
is a tensor of order $d$ with $P^d$ coefficients.
Usually, such tensors cannot be computed exactly and explicitly 
with $d \geq 2$ (which includes the Hessian) for medium-sized
models ($P \gtrsim 10^6$).

Instead of trying to approximate these tensors,
we propose to compute their projections along well-chosen directions,
that are relevant for optimization.
This computation can be done efficiently by taking advantage of
the practical implementation of the vector of parameters 
$\btheta$ as a tuple of tensors $(\bT^1, \cdots, \bT^S)$.
Such a projection of the order-$d$ derivative
yields a tensor of order $d$ with $S^d$ coefficients, 
instead of $P^d$. 
Thus, the Hessian of a model of size $P = 10^6$ 
represented by a tuple of $S = 20$ tensors
can be reduced to a matrix of size $S^2 = 400$,
instead of $P^2 = 10^{12}$.
More generally, whenever $S \ll P$, 
the projected order-$d$ derivative of $\caL$ is significantly smaller and easier to compute than 
the full order-$d$ derivative.

\paragraph{Application: computing per-layer learning rates.}
Then, we show that such projections of the order-1 and order-2 derivatives 
can be used to compute the optimal learning rates to choose for each
one of the $S$ tensors (or subsets of parameters).
The procedure we propose to compute per-layer learning rates 
is both theoretically well-grounded and usable in practice (as long as
the number of layers is not too large).
In particular, our computation does not neglect long-range interactions between layers.

\paragraph{Application: second-order optimization method.}
Finally, we show that the information contained in the 1st, 2nd, and 3rd order 
derivatives is not only accessible at reasonable cost, but can also be used for optimization. 
In particular, we propose an optimization method that exploits 
higher-order information about the loss obtained by using the main contribution. 
For simplicity, our optimization method and Newton's method look similar: 
in both cases, a linear system $\bH_0 \mathbf{x} = \bg_0$ has to be solved 
(w.r.t.\ $\mathbf{x}$), where $\bg_0$ and $\bH_0$ contain 
respectively first-order and second-order information about $\caL$.
Despite this formal resemblance, the difference is enormous:
with Newton's method, $\bH_0$ is equal to the Hessian $\bH$ of $\caL$ of size $P \times P$,
while with ours, $\bH_0$ is equal to a matrix $\bar{\bH}$ of size $S \times S$.
Thus, $\bar{\bH}$ is undoubtedly smaller and easier to compute than $\bH$
when $S \ll P$. Nevertheless, since $\bar{\bH}$ is a dense matrix,
it still contains information about the interactions between the tensors $\bT^s$
when they are used in $\caL$.
This point is crucial because most second-order optimization methods
applied to neural networks use a simplified version of the Hessian
(or its inverse), usually a diagonal or block-diagonal approximation,
ignoring interactions between layers.
Additionally, we propose an anisotropic version of Nesterov's cubic regularization
\cite{nesterov2006cubic}, which uses order-3 information to regularize $\bar{\bH}$ and avoid 
instabilities when computing $\bar{\bH}^{-1} \bar{\bg}$.
In particular, the resulting training trajectory is invariant by layer-wise affine reparameterizations,
so our method preserves some interesting properties of Newton's method.

\paragraph{Structure of the paper.}
First, we show the context and motivation of our work in Section \ref{sec:context}.
Then, we provide in Section \ref{sec:short:higher_order} our core method,
and in Sections \ref{sec:optim:presentation} and \ref{sec:short:optim} 
its applications.
In Section \ref{sec:expes}, we present experimental results
showing that the developed methods are usable in practice.
Finally, we discuss the results in Section \ref{sec:discussion}.

\section{Context and motivation} \label{sec:context}

\subsection{Higher-order information}

It is not a novel idea to 
extract higher-order information about a loss
at a minimal computational cost to improve optimization.
This is typically what is done by \cite{dangel2023backpropagation},
although it does not go beyond the second-order derivative.
In this line of research, the \emph{Hessian-vector product}
\cite{pearlmutter1994fast}
is a decisive tool, that allows to compute
the projection of higher-order derivatives in given directions 
at low cost (see App.~\ref{app:fast_comp}).
For derivatives of order $3$, 
Nesterov's cubic regularization of Newton's method 
\cite{nesterov2006cubic} uses information of order $3$
to avoid too large training steps. Incidentally,
we develop an anisotropic variant of this in Section \ref{sec:short:optim}.
In the same spirit, the use of derivatives of any order for optimization
has been proposed \cite{birgin2017worst}

\subsection{Using and estimating the Hessian in optimization} \label{sec:optim:context}

The Hessian $\bH$ of the loss $\caL$ according to the vector 
of parameters $\btheta$ is known to contain useful information about $\caL$.
Above all, the Hessian is used to develop second-order optimization
algorithms. Let us denote by $\btheta_t$ the value of $\btheta$ at time step $t$,
$\bg_t \in \mathbb{R}^P$ the gradient of $\caL$ at step $t$ and $\bH_t$
its Hessian at step $t$.
One of the most widely known second-order optimization method is Newton's method,
whose step is \cite[Chap.\ 3.3]{nocedal1999numerical}:
\begin{align}
\btheta_{t + 1} := \btheta_t - \bH_t^{-1} \bg_t .
\end{align}
Under certain conditions, including strong convexity of $\caL$,
the convergence rate of Newton's method is quadratic
\cite[Th.\ 3.7]{nocedal1999numerical}, which makes it
very appealing.
Besides, other methods use second-order information
without requiring the full computation of the Hessian.
For instance, Cauchy's steepest descent \cite{cauchy1847methode} is 
a variation of the usual gradient descent, where the step size is
tuned by extracting very little information from the Hessian:
\begin{align}
\btheta_{t + 1} := \btheta_t - \eta_t^* \bg_t,
\quad \text{where} \quad \eta_t^* := \frac{\bg_t^T \bg_t}{\bg_t^T \bH_t \bg_t} ,
\end{align}
where the value of $\bg_t^T \bH_t \bg_t$ can be obtained with little computational cost
(see Appendix \ref{app:fast_comp}).
However, when optimizing a quadratic function $f$ with Cauchy's steepest descent,
$f(\theta_t)$ is known to decrease at a rate $(\frac{\lambda_{\mathrm{max}} - \lambda_{\mathrm{min}}}{
\lambda_{\mathrm{max}} + \lambda_{\mathrm{min}}})^2$, where $\lambda_{\mathrm{max}}$ and $\lambda_{\mathrm{min}}$
are respectively the largest and the smallest eigenvalues of the Hessian of $f$ 
\cite[Chap.\ 8.2, Th.\ 2]{luenberger2008linear}.
If the Hessian of $f$ is
strongly anisotropic, then this rate is close to one and optimization is slow.
For a comparison of the two methods, see
\cite{gill1981practical,luenberger2008linear,nocedal1999numerical}.

Finally, there should be some space between Newton's method, 
which requires the full Hessian $\bH$,
and Cauchy's steepest descent, which requires minimal and computationally cheap information
about $\bH$. 
The optimization method presented in Section \ref{sec:short:optim}
explores this in-between space.

\paragraph{Quasi-Newton methods.}
When the parameter space is high-dimensional, 
computation of the Hessian $\bH_t$ and
inversion of the linear system $\bg_t = \bH_t \mathbf{x}$
are computationally intensive.
Quasi-Newton methods are designed to avoid any direct computation of the Hessian,
and make extensive use of gradients and finite difference methods
to approximate the direction of $\bH_t^{-1} \bg_t$.
For a list of quasi-Newton methods, see
\cite[Chap.\ 8]{nocedal1999numerical}.
However, \cite{nocedal1999numerical} argue that,
since it is easy to compute the Hessian
by using Automatic Differentiation (AutoDiff), quasi-Newton methods tend to lose their interest. 

\paragraph{Applications to deep learning.}
Many methods overcome the curse of the number of parameters by exploiting 
the structure of the neural networks. It is then common to
neglect interactions between layers, leading to a (block)-diagonal 
approximation of the Hessian.
A first attempt has been made by \cite{wang1998second}:
they divide the Hessian into blocks, 
following the division of the network into layers, 
and its off-diagonal blocks are removed.
From another perspective, \cite{ollivier2015riemannian}
keeps this block-diagonal structure, but performs an additional
approximation on the remaining blocks. 

More recently, K-BFGS has been proposed \cite{goldfarb2020practical},
which is a variation of the quasi-Newton method BFGS with block-diagonal 
approximation and an approximate representation of these blocks.
In a similar spirit, the Natural Gradient method TNT \cite{ren2021tensor}
also exploits the structure of neural networks
by performing a block-diagonal approximation. 
Finally, AdaHessian \cite{yao2021adahessian} efficiently implements
a second-order method by approximating the Hessian by its diagonal.

\emph{Kronecker-Factored Approximate Curvature} (K-FAC)
is a method for approximating of the Hessian proposed in 
\cite{martens2015optimizing} in the context of neural network training.
K-FAC exploits the specific architecture of neural networks
to output a cheap approximation of the true Hessian.
Despite its scalability, K-FAC suffers from several problems.
First, the main approximation is quite rough, since 
``[it assumes] statistical independence between products [...] of unit activities 
and products [...] of unit input derivatives'' \cite[Sec.\ 3.1]{martens2015optimizing}.
Second, even with an approximation of the Hessian, one
has to invert it, which is computationally intensive
even for small networks.
To overcome this difficulty,
a block-(tri)diagonal approximation of the inverse of
the Hessian is made, which eliminates many of the interactions between the layers. 

\paragraph{Summarizing the Hessian.}
In Section \ref{sec:short:optim}, we propose to summarize the Hessian to 
avoid the expensive computation of the full Hessian. 
This idea is not new. For instance, \cite{lu2018block} proposes
to approximate the Hessian with a matrix composed of
blocks in which all coefficients are identical.
A more broadly used technique to compress the Hessian
is to perform \emph{sketching} on it, that is, project it
on randomly chosen directions. This idea
is used for solving solve linear systems \cite{yuan2022sketched},
as well as for minimizing functions \cite{gower2019rsn},
and can be further adapted to Newton's method 
with cubic regularization \cite{hanzely2020stochastic}.
Finally, it is also possible to choose the directions of the projection 
by using available information \cite{nonomura2021randomized}. 
This is the strategy that we have adopted in Section \ref{sec:short:optim}.

\paragraph{Invariance by affine reparameterization.}
Several optimization methods,
such as Newton's, have an optimization step invariant by
affine reparameterization of $\btheta$ 
\cite{amari1998natural} \cite[Chap.\ 4.1.2]{nesterov2003introductory}.
Specifically, when using Newton's method,
it is equivalent to optimize $\caL$ according to $\btheta$ and according to 
$\tilde{\btheta} = \mathbf{A} \btheta + \mathbf{B}$ ($\mathbf{A} \in \mathbb{R}^{P \times P}$
invertible, $\mathbf{B} \in \mathbb{R}^{P}$).
This affine-invariance property holds even if the function $\caL$ to minimize
is a negative $\log$-likelihood, and one chooses to minimize $\btheta$ 
by the \emph{natural gradient} method \cite{amari1998natural}.
This method also requires computing the Hessian of $\caL$ at some point.

\paragraph{Methods based on the moments of the gradients.}
Finally, many methods acquire geometric information on the loss
by using only the gradients. For instance, Shampoo \cite{gupta2018shampoo}
uses second-moment information of the accumulated gradients.

\subsection{Motivation}

\paragraph{What are we really looking for?}
The methods that aim to estimate the Hessian matrix $\bH$
or its inverse $\bH^{-1}$ in order to imitate Newton's method implicitly
assume that Newton's method is adapted to the current problem.
This assumption is certainly correct when the loss to optimize is strongly convex.
But, when the loss is not convex and very complicated, e.g.\ when training a 
neural network, this assumption is not justified.
Worse, it has been shown empirically that, at the end of the training of a neural network,
the eigenvalues of the Hessian are concentrated around zero
\cite{sagun2018empirical}, with only a few large positive eigenvalues.
Therefore, Newton's method itself does not seem to be recommended for neural network
training, so we may not need to compute the full Hessian at all,
which would relieve us of a tedious, if not impossible, task.

To avoid such problems, it is very common to regularize the Hessian
by adding a small, constant term $\lambda \bI$ to it 
\cite[Chap.\ 6.3]{nocedal1999numerical}. Also, trust-region Newton 
methods are designed to handle non-positive-definite Hessian matrices
\cite[Chap.\ 6.4]{nocedal1999numerical} \cite{nash1984newton}.

\paragraph{Importance of the interactions between layers.}
Also, some empirical works have shown that the role and the behavior
of each layer must be considered along its interactions with the other layers,
which emphasize the importance of 
off-diagonal blocks in the Hessian or its inverse.
We give two examples. First, \cite{zhang2022all} has shown that,
at the end of their training, many networks exhibit a strange
feature: some (but not all) layers can be reinitialized to their initial value
with little loss of the performance.
Second, \cite{kornblith2019similarity} has compared the similarity between 
the representations of the data after each layer:
changing the number of layers can qualitatively change
the similarity matrix of the layers 
\cite[Fig. 3]{kornblith2019similarity}.
Among all, these results motivate our search for 
mathematical objects that show how layers interact.

\paragraph{Per-layer scaling of the learning rates.}
A whole line of research is concerned with building a well-founded method
for finding a good scaling for the initialization distribution of the parameters, 
and for the learning rates, which can be chosen layer-wise.
For instance, a layer-wise scaling for the weights
was proposed and theoretically justified in the paper introducing the Neural Tangent 
Kernels \cite{jacot2018neural}.
Also, in the ``feature learning'' line of work, 
\cite{yang2021tensor} proposes a relationship between different
scalings related to weight initialization and training.
Therefore, there is an interest in finding a scalable and theoretically grounded method to 
build per-layer learning rates.

\paragraph{Unleashing the power of AutoDiff.}
Nowadays, several libraries provide easy-to-use automatic 
differentiation packages that allow the user to compute
numerically the gradient of a function, and even higher-order derivatives.%
\footnote{With PyTorch: torch.autograd.grad.}
Ignoring the computational cost, the full Hessian could theoretically
be computed numerically without any approximation.
To make this computation feasible, one should
aim for an simpler goal: instead of computing the Hessian, one can 
consider a smaller matrix, consisting of projections of the Hessian.

Moreover, one might hope that such projections would ``squeeze'' the close-to-zero
eigenvalues of the Hessian, so that the eigenvalues of the projected matrix
would not be too close to zero.

\section{Summarizing higher-order information} \label{sec:short:higher_order}

Let us consider the minimization of a loss function $\caL : \mathbb{R}^P \rightarrow \mathbb{R}$
according to a variable $\btheta \in \mathbb{R}^P$. 

\paragraph{Full computation of the derivatives.}
The order-$d$ derivative of $\caL$ at a point $\btheta$, that we denote by
$\frac{\dd^d \caL}{\dd \btheta^d}(\btheta)$, can be viewed as either
a $d$-linear form (see \cite{dieudonne_foundations_1960} and Appendix 
\ref{app:dieudonne}) or as an order-$d$ tensor belonging to $\mathbb{R}^{P^d}$. For convenience, we will
use the latter: the coefficients of the tensor $\mathbf{A} = \frac{\dd^d \caL}{\dd \btheta^d}(\btheta) 
\in \mathbb{R}^{P^d}$ are $A_{i_1, \cdots, i_d} = \frac{\partial^d \caL}{
	\partial \theta_{i_1} \cdots \partial \theta_{i_d}}(\btheta)$,
where $(i_1, \cdots, i_d) \in \{1, \cdots, P\}^d$ is a multi-index.
For a tensor $\mathbf{A} \in \mathbb{R}^{P^d}$,
we will use the following notation for tensor contraction:
\begin{align}
	\forall (\bu_1, \cdots, \bu_d) \in \mathbb{R}^P \times \cdots \times 
	\mathbb{R}^P, \nonumber \\
	\mathbf{A}[\bu^1, \cdots, \bu^d]
	:= \sum_{i_1 = 1}^{P} \cdots \sum_{i_d = 1}^{P} A_{i_1, \cdots, i_d} 
	u^1_{i_1} \cdots u^d_{i_d}.
\end{align}
The order-$d$ derivative $\frac{\dd^d \caL}{\dd \btheta^d}(\btheta) \in \mathbb{R}^{P^d}$
contains $P^d$ scalars.
But, even when considering its symmetries, 
it is computationally too expensive to compute it exactly for $d \geq 2$ in most cases.
For instance, it is not even possible to compute numerically the full Hessian of $\caL$
according to the parameters of a small neural network, i.e., 
with $P = 10^5$ and $d = 2$, the Hessian contains $P^d = 10^{10}$ scalars.

\paragraph{Terms of the Taylor expansion.}
At the opposite, one can obtain cheap higher-order information
about $\caL$ at $\btheta$ by considering a specific direction $\bu \in \mathbb{R}^P$.
The Taylor expansion of $\caL(\btheta + \bu)$ gives:
\begin{align}
	\caL(\btheta + \bu) = \caL(\btheta) + \sum_{d = 1}^{D} \frac{1}{d!}
	\frac{\dd^d \caL}{\dd \btheta^d}(\btheta)[\bu, \cdots , \bu] 
	+ o(\|\bu\|^D) .
\end{align}
The terms of the Taylor expansion contain higher-order information
about $\caL$ in the direction $\bu$. In particular, they can be used to 
predict how $\caL(\btheta)$ would change if $\btheta$ was translated
in the direction of $\bu$.
Additionally, computing the first $D$ terms has a complexity of order
$D \times P$, which is manageable even for large models.
The trick that allows for such a low complexity, the \emph{Hessian-vector product},
was proposed by 
\cite{pearlmutter1994fast} and is recalled in Appendix \ref{app:fast_comp}.

\paragraph{An intermediate solution.}
Now, let us assume that, in the practical implementation of a gradient-based
method of optimization of $\mathcal{L}(\btheta)$,
$\btheta$ is represented by a tuple of tensors $(\bT^1, \cdots, \bT^S)$.
So, each Taylor term can be expressed as:
\begin{align}
	\frac{\dd^d \caL}{\dd \btheta^d}(\btheta)[\bu, \cdot \,\cdot , \bu] 
	&= \sum_{s_1 = 1}^{S} \cdots \sum_{s_d = 1}^{S}
	\frac{\partial^d \caL}{\partial \bT^{s_1} \cdots \partial \bT^{s_d}}(\btheta)
	[\bU^{s_1}, \cdots , \bU^{s_d}] \nonumber \\
	&= \mathbf{D}_{\btheta}^{d}(\bu)[\mathds{1}_S, \cdots, \mathds{1}_S] , \label{eqn:taylor}
\end{align}
where $\mathds{1}_S \in \mathbb{R}^S$ is a vector full of ones,
the tuple of tensors $(\bU^{1}, \cdots, \bU^{S})$ represents $\bu$,%
\footnote{$(\bU^1, \cdots, \bU^S)$ is to $\bu$ 
	as $(\bT^1, \cdots,  \bT^S)$ is to $\btheta$.}
and $\mathbf{D}_{\btheta}^{d}(\bu) \in \mathbb{R}^{S^d}$ is a
tensor of order $d$ with size $S$ in every dimension s.t.:
\begin{align}
	(\mathbf{D}_{\btheta}^{d}(\bu))_{s_1, \cdots, s_d}
	&= \frac{\partial^d \caL}{\partial \bT^{s_1} \cdots \partial \bT^{s_d}}(\btheta)
	[\bU^{s_1}, \cdots , \bU^{s_d}] \label{eqn:defD}\\
	&= \sum_{i_1 = 1}^{P_{s_1}} \cdots \sum_{i_1 = 1}^{P_{s_d}}
	\frac{\partial^d \caL}{\partial T^{s_1}_{i_1} \cdots \partial T^{s_d}_{i_d}}(\btheta)
	U^{s_1}_{i_1} \cdots U^{s_d}_{i_d} ,
\end{align}
where $P_s$ is the number of coefficients of the tensor $\bT^s$.
Thus, $\mathbf{D}_{\btheta}^{d}(\bu)$ is 
a tensor of order $d$ and size $S$ in every dimension
resulting from a partial contraction of the full derivative
$\frac{\dd^d \caL}{\dd \btheta^d}(\btheta)$.
Moreover, the trick of \cite{pearlmutter1994fast} also applies to
the computation of $\mathbf{D}_{\btheta}^{d}(\bu)$, which is then much less expensive to
compute than the full derivative (see Appendix \ref{app:fast_comp}).

\paragraph{Properties of $\mathbf{D}_{\btheta}^{d}(\bu)$.}
We show a comparison between the three techniques in Table \ref{tbl:comparison}.
If $S$ is small enough, computing $\mathbf{D}_{\btheta}^{d}(\bu)$ becomes 
feasible for $d \geq 2$.
For usual multilayer perceptrons with $L$ layers, 
there is one tensor of weights and one vector of biases per layer,
so $S = 2 L$.
This allows to compute $\mathbf{D}_{\btheta}^{d}(\bu)$ in practice
for $d = 2$ even when $L \approx 20$.

\begin{table}[t]
	\caption{Comparison between three techniques extracting higher-order
		information about $\caL$:
		size of the result and complexity of the computation.} 
	\label{tbl:comparison}
	\vspace*{3mm}
	\centering
	\begin{tabular}{lcc}
		\toprule
		Technique & Size & Complexity \\ \midrule
		Full derivative $\frac{\dd^d \caL}{\dd \btheta^d}(\btheta)$
		& $P^d$ & $P^d$ \\
		Taylor term $\mathbf{D}_{\btheta}^{d}(\bu)[\mathds{1}_S, \cdots, \mathds{1}_S]$
		& $1$ & $d \times P$ \\
		Tensor $\mathbf{D}_{\btheta}^{d}(\bu)$ 
		& $S^d$ & $S^{d - 1} \times P$ \\
		\bottomrule
	\end{tabular}
\end{table}

{According to Eqn.~\eqref{eqn:taylor}, the Taylor term can be obtained by
full contraction of $\mathbf{D}_{\btheta}^{d}(\bu)$. However, $\mathbf{D}_{\btheta}^{d}(\bu)$,
is a tensor of size $S^d$, and cannot be obtained from the Taylor term, which is only a scalar.
Thus, the tensors $\mathbf{D}_{\btheta}^{d}(\bu)$ extract
more information than the Taylor terms, while keeping a reasonable computational cost.
Moreover, their off-diagonal elements give access to information about one-to-one interactions between 
tensors $(\bT^1, \cdots, \bT^S)$ when they are processed in the function $\caL$.

\section{Application: computing per-layer learning rates} \label{sec:optim:presentation}

To build per-layer (or per-subset-of-parameters) learning rates,
we partition the set of indices of parameters
$\{1, \cdots, P\}$ into $S$ subsets $(\mathcal{I}_s)_{1 \leq s \leq S}$,
we assign for all $1 \leq s \leq S$ the same learning rate $\eta_s$ to
the parameters $(\theta_p)_{p \in \mathcal{I}_s}$,
and we find the vector of learning rates 
$\bfeta = (\eta_1, \cdots , \eta_S)$
optimizing the decrease of the loss $\caL$
for the current training step $t$, by using its order-2 Taylor approximation.%
\footnote{With the notation of Section \ref{sec:short:higher_order}, 
$\mathcal{I}_s$ is the set of indices $p$ of the parameters $\theta_p$
belonging to the tensor $\bT^s$, so 
the scalars $(\theta_p)_{p \in \mathcal{I}_s}$ correspond
to the scalars belonging to $\bT^s$. So, everything is as if
a specific learning rate $\eta_s$ is assigned to each $\bT^s$.}
Formally, given a direction $\bu_t \in \mathbb{R}^P$ in the
parameter space (typically, $\bu_t =  \bg_t$, the gradient)
and $\bU_t := \mathrm{Diag}(\bu_t) \in \mathbb{R}^{P \times P}$,
we consider the training step:
$\btheta_{t + 1} := \btheta_t - \bU_t \bI_{P:S} \bfeta_t$,
that is a training step in a direction based on $\bu_t$, 
distorted by a subset-wise step size $\bfeta_t$. 
Then, we
minimize the order-2 Taylor approximation of 
$\caL(\btheta_{t + 1}) - \caL(\btheta_t)$: 
$\bDelta_2(\bfeta_t) := -\bg_t^T \bU_t \bI_{P:S} \bfeta_t
	+ \frac{1}{2} \bfeta_t^T \bI_{S:P} \bU_t \bH_t \bU_t \bI_{P:S} \bfeta_t$,
which gives:
\begin{align}
	\btheta_{t + 1} &= \btheta_t - \bU_t \bI_{P:S} \bfeta_t^* ,  &
	\bfeta_t^* &:= 
	(\bI_{S:P} \bU_t \bH_t \bU_t \bI_{P:S})^{-1} \bI_{S:P} \bU_t \bg_t , \label{eqn:ours}
\end{align}
where $\bI_{S:P} \in \mathbb{R}^{S \times P}$ is the \emph{partition matrix},
verifying $(\bI_{S:P})_{sp} = 1$ if $p \in \mathcal{I}_s$ and
$0$ otherwise, and $\bI_{P:S} := \bI_{S:P}^T$. Alternatively, $\bfeta_t^*$
can be written (details are provided in Appendix \ref{app:derivation}):
\begin{align}
	\bfeta_t^* &= \bar{\bH}_t^{-1} \bar{\bg}_t, & 
	\text{where:} \quad \bar{\bH}_t := \bI_{S:P} \bU_t \bH_t \bU_t \bI_{P:S} 
	\in \mathbb{R}^{S \times S}, \quad 
	\bar{\bg}_t := \bI_{S:P} \bU_t \bg_t \in \mathbb{R}^S .
\end{align}

With the notation of Section \ref{sec:short:higher_order}, 
$\bar{\bH}_t = \mathbf{D}_{\btheta_t}^{(2)}(\bu_t)$
and $\bar{\bg}_t = \mathbf{D}_{\btheta_t}^{(1)}(\bu_t)$.
Incidentally, computing $\bar{\bH}$ is of complexity
$S P$, and solving $\bar{\bH} \mathbf{x} = \bar{\bg}$
is of complexity $S^2$.

\section{Application: optimization method} \label{sec:short:optim}

\subsection{Presentation}

Now that we can can compute per-layer learning rates, we
decide to incorporate them into an optimization method.
However, computing them requires to compute $\bar{\bH}^{-1} \bar{\bg}$.
Usually, inverting such a linear system at every step is considered as hazardous
and unstable. Therefore, when using Newton's method, instead of computing
descent direction $\bu := \bH^{-1} \bg$, 
it is very common to add a regularization term:
$\bu_{\lambda} := \left(\bH + \lambda \bI\right)^{-1} \bg$
\cite[Chap.\ 6.3]{nocedal1999numerical}.

However, the theoretical ground of such a regularization technique
is not fully satisfactory.
Basically, the main problem is not having a matrix $\bar{\bH}$ with close-to-zero eigenvalues:
after all, if the loss landscape is very flat in a specific direction, 
it is better to make a large training step. 
The problem lies in the order-2 approximation of the loss
made in the training step \eqref{eqn:ours}, as well as in Newton's method:
instead of optimizing the true decrease of the loss, we optimize the decrease
of its order-2 approximation.
Thus, the practical question is: does this approximation faithfully model
the loss at the current point $\btheta_t$, in a region that also includes
the next point $\btheta_{t + 1}$?

To answer this question, one has to take into account order-3 information,
and regularize $\bar{\bH}$ so that the resulting update remains
in a region around $\btheta_t$ where the cubic term of the Taylor approximation
is negligible.
In practice, we propose an anisotropic version of Nesterov's cubic regularization
\cite{nesterov2006cubic}.

\paragraph{Anisotropic Nesterov cubic regularization.}
By using the technique presented in Section \ref{sec:short:higher_order},
the diagonal coefficients $(D_1, \cdots, D_S)$
of $\mathbf{D}_{\btheta}^{(3)}(\bu) 
\in \mathbb{R}^{S \times S \times S}$
are available with little computational cost.
Let
$\bD := \mathrm{Diag}(|D_1|^{1/3}, \cdots, |D_S|^{1/3}) \in \mathbb{R}^S$.

We modify the method of \cite{nesterov2006cubic} by 
integrating an anisotropic factor $\mathbf{D}$ into the cubic term.
Thus, our goal is to minimize according to $\bfeta$ the function $T$:
$T(\bfeta) := - \bfeta^T \bar{\bg} + \frac{1}{2} \bfeta \bar{\bH} \bfeta 
+ \frac{\lambda_{\mathrm{int}}}{6} \|\bD \bfeta\|^3$, 
where $\lambda_{\mathrm{int}}$ is the \emph{internal damping}
coefficient, which can be used to tune the strength of the cubic
regularization. Under conditions detailed in Appendix \ref{app:nesterov},
this minimization problem is equivalent to finding a solution $\bfeta_*$ such that:
\begin{align}
\bfeta_* =
\left(\bar{\bH} + \frac{\lambda_{\mathrm{int}}}{2} \|\bD \bfeta_*\| \bD^2\right)^{-1}\bar{\bg},
\label{eqn:nesterov}
\end{align}
which is a regularized version of \eqref{eqn:ours}.
Finally, this multi-dimensional minimization 
problem boils down to a scalar root finding problem (see Appendix \ref{app:nesterov}).

\subsection{Properties}

The final method is a combination of the learning rate computed in Eqn.~\eqref{eqn:ours} 
with regularization \eqref{eqn:nesterov}:
\begin{method} \label{meth:order2}
	Training step $\btheta_{t + 1} = \btheta_t - \bU_t \bI_{P:S} \bfeta_t^*$, where
	$\bfeta_t^*$ is the solution with the largest norm $\|\bD_{t} \bfeta\|$ of the equation:
	$\bfeta =
		\left(\bar{\bH}_{t} + \frac{\lambda_{\mathrm{int}}}{2} \|\bD_{t} \bfeta\| 
		\bD_{t}^2\right)^{-1}\bar{\bg}_{t}$.
\end{method}

\paragraph{Encompassing Newton's method and Cauchy's steepest descent.}
Without the cubic regularization ($\lambda_{\mathrm{int}} = 0$), 
Newton's method is recovered when using the \emph{discrete partition}, that
is, $S = P$ with $\mathcal{I}_s = \{s\}$ for all $s$,
and Cauchy's steepest descent is recovered when using
the \emph{trivial partition}, that is, $S = 1$ with 
$\mathcal{I}_1 = \{1, \cdots , P\}$. See Appendix \ref{app:link_cauchy}
for more details.

\paragraph{No need to compute or approximate the full Hessian.}
The full computation of the Hessian $\bH_t \in \mathbb{R}^{P \times P}$ is not required. 
Instead, one only needs to compute the $S \times S$ matrix
$\metH_t := \bI_{S:P} \bU_t \bH_t \bU_t \bI_{P:S}$,
which can be done efficiently by computing 
$\bu^T \bH_t \bv$ for a number $S \times S$ of pairs of well-chosen directions 
$(\bu, \bv) \in \mathbb{R}^P \times \mathbb{R}^P$. 
This property is especially useful when $S \ll P$.
When optimizing a neural network with $L = 10$ 
layers and $P = 10^6$ parameters, one can naturally partition
the set of parameters into $S = 2 L$ subsets,
each one containing either all the weights or all the biases of each of the
$L$ layers. In this situation, one has to solve a linear system 
of size $2 L = 20$ at each step, which is much more reasonable
than solving a linear system of $P = 10^6$ equations.
We call this natural partition of the parameters of a neural network
the \emph{canonical partition}.

\paragraph{No need to solve a large linear system.}
Using Equations \eqref{eqn:ours} or \eqref{eqn:nesterov} requires solving only
a linear system of $S$ equations, instead of $P$ in Newton's method.
With the cubic regularization, only a constant term is added to the complexity, 
since it is a matter of scalar root finding.

\paragraph{The interactions between different tensors are not neglected.}
The matrix $\metH_t$, which simulates the Hessian $\bH_t$,
is basically dense:
it does not exhibit a (block-)diagonal structure.
So, the interactions between subsets of parameters are taken into account
when performing optimization steps.
In the context of neural networks with the canonical partition, 
this means that interactions between layers are taken into account
during optimization, even if the layers are far from
each other. This is a major advantage over
many existing approximations of the Hessian or its inverse, 
which are diagonal or block-diagonal.

\paragraph{Invariance by subset-wise affine reparameterization.}
As showed in Appendix \ref{app:invariance}, 
under a condition on the directions $\bu_t$,%
\footnote{It holds if $\bu_t$ is the gradient or
	a moving average of the gradients (momentum).}
the trajectory of optimization of a model trained by Method \ref{meth:order2}
is invariant by affine reparameterization
of the sub-vectors of parameters $\btheta_{\mathcal{I}_s}
:= \mathrm{vec}(\{\theta_p : p \in \mathcal{I}_s\})$.
Let $(\alpha_s)_{1 \leq s \leq S}$ and
$(\beta_s)_{1 \leq s \leq S}$ be a sequence of nonzero scalings
and a sequence of offsets, and $\tilde{\btheta}$ such that, 
for all $1 \leq s \leq S$, $\tilde{\btheta}_{\mathcal{I}_s} = \alpha_s \btheta_{\mathcal{I}_s} + \beta_s$.
Then, the training trajectory of the model is the same with both parameterizations 
$\btheta$ and $\tilde{\btheta}$.
This property is desirable in the case of neural networks,
where one can use either the usual or
the NTK parameterization, which consists of a layer-wise scaling of
the parameters. The relevance of this property is discussed
in Appendix \ref{app:ntk}.

Compared to the standard regularization $\bar{\bH} + \lambda \bI$ and
Nesterov's cubic regularization, the anisotropic Nesterov regularization
does not break the property of invariance by subset-wise scaling of the parameters
of \eqref{eqn:ours}.
This is mainly due to our choice to keep only the diagonal coefficients of 
$\mathbf{D}_{\btheta}^{(3)}(\bu)$ while discarding the others. 
In particular, the off-diagonal coefficients contain cross-derivatives
that would be difficult to include in an invariant training step.

\section{Experiments} \label{sec:expes}

\subsection[Empirical computation of H and eta]{Empirical computation of $\bar{\bH}$ and $\bfeta_*$}

As recalled in Section \ref{sec:context}, many works
perform a diagonal, block-diagonal or block-tridiagional \cite{martens2015optimizing}
approximation of the Hessian or its inverse.
Since a summary $\bar{\bH}$ of the Hessian and its inverse $\bar{\bH}^{-1}$ are available
and all their off-diagonal coefficients have been computed and kept, one can
to check if these coefficients are indeed negligible.

\paragraph{Setup.}
We have trained LeNet-5 and VGG-11'%
\footnote{VGG-11' is a variant of VGG-11 with 1 final fully-connected
	layer instead of 3.}
on CIFAR-10 using SGD with momentum.
Before each epoch, we compute the full-batch gradient, denoted by $\bu$,
which we use as a direction to compute $\bar{\bH}$, again in full-batch.
We report submatrices of $\bar{\bH}$ and $\bar{\bH}^{-1}$ at initialization
and at the epoch where the validation loss is the best in 
Figure \ref{fig:H:lenet} (LeNet) and Figure \ref{fig:H:vgg} (VGG-11').

For the sake of readability, 
$\bar{\bH}$ has been divided into blocks: 
a weight-weight block $\bar{\bH}_{\text{WW}}$, 
a bias-bias block $\bar{\bH}_{\text{BB}}$,
and a weight-bias block $\bar{\bH}_{\text{WB}}$.
They represent the interactions between the layers:
for instance, $(\bar{\bH}_{\text{WB}})_{l_1 l_2}$ 
represents the interaction between the tensor of weights of layer $l_1$
and the vector of biases of layer $l_2$. 

\paragraph{Results on $\bar{\bH}$.}
First, the block-diagonal
approximation of the Hessian is indeed very rough,
while the block-diagonal approximation of the inverse Hessian
seems to be more reasonable (at least in these setups), which has already been
shown by \cite{martens2015optimizing}.
Second, there seem to be long-range interactions between layers, both at initialization
and after several epochs. For LeNet, all the layers (except the first one)
seem to interact together at initialization (Fig.\ \ref{fig:H:lenet}).
In the matrix $\bar{\bH}^{-1}$ computed on VGG, the last 3 layers
interact strongly and the last 6 layers also interact, but a bit less.

According to these observations, a neural network should also be considered as a whole,
in which layers can hardly be studied independently from each other.
To our knowledge, this result is the first scalable representation of
interactions between distant layers, based on 
second-order information.

\begin{figure}[ht]
	\begin{subfigure}{.48\linewidth}
		\includegraphics[width=\linewidth]{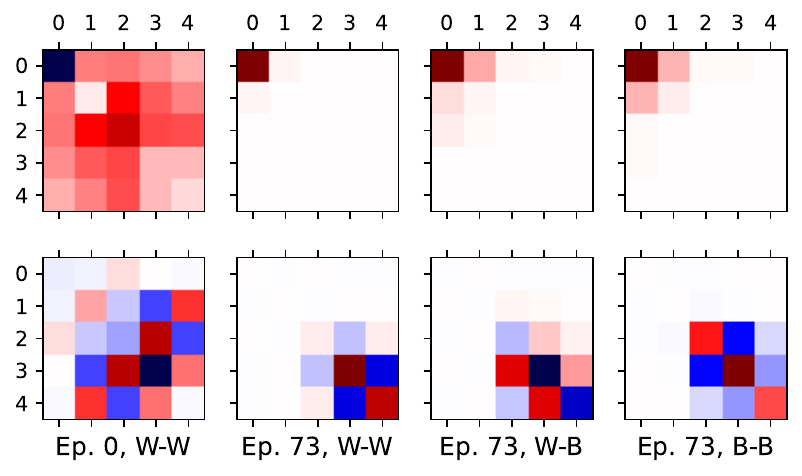}
		\subcaption{LeNet + CIFAR10.}
		\label{fig:H:lenet}
	\end{subfigure}
	$\;$
	\begin{subfigure}{.48\linewidth}
		\includegraphics[width=\linewidth]{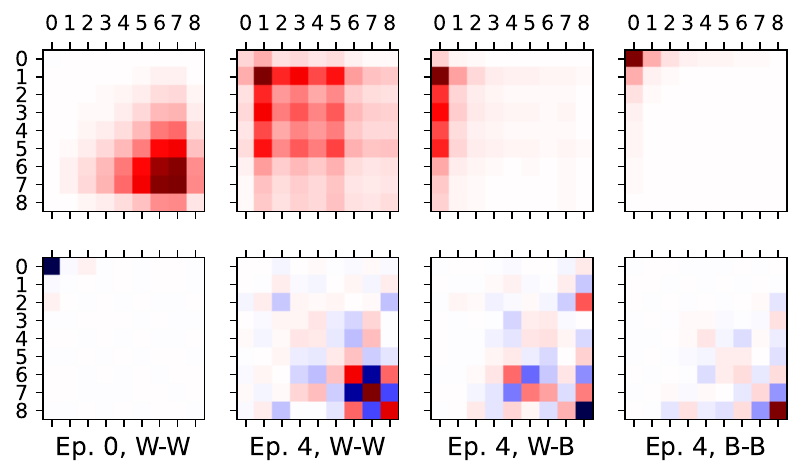}
		\subcaption{VGG-11' + CIFAR10.}
		\label{fig:H:vgg}
	\end{subfigure}
	\caption{Setup: models trained by SGD on CIFAR-10.
		Submatrices of $\bar{\bH}$ (1st row) and $\bar{\bH}^{-1}$ (2nd row),
		where focus is on interactions: weight-weight, weight-bias, bias-bias of 
		the different layers, at initialization and before best validation loss.
	}
	\label{fig:H}
\end{figure}

\paragraph{Results on $\bfeta_*$.}
The evolution of the learning rates $\eta_*$ computed according to \eqref{eqn:nesterov}
in LeNet and VGG is shown in Figure \ref{fig:lrs}. 
First, the learning rates computed
for the biases are larger than those computed for the weights.
Second, even if only the weights are considered, the computed $\eta_*$
can differ by several orders of magnitude. 
Finally, the first two layers of LeNet (which are convolutional)
have smaller $\eta_*$ than the last three layers (which are fully-connected).
Conversely, in VGG, the weights of the last (convolutional) layers have 
a smaller $\eta_*$ than those of the first layers.

\subsection{Training experiments} \label{sec:training}

To show that the projections of the 2nd and 3rd order derivatives of the loss 
defined in Section \ref{sec:short:higher_order} can be practically used to 
train neural networks, we test our optimization method
\ref{meth:order2} (summarized in Algorithm \ref{alg:main_reduc}) on simple vision tasks.
All the implementation details are available in Appendix \ref{app:implementation}.
In particular, we have introduced a step size $\lambda_1$ that leads to the
following modification of the training step \eqref{eqn:ours}:
$\btheta_{t + 1} = \btheta_t - \lambda_1 \bU_t \bI_{P:S} \bfeta_t^*$.

\begin{algorithm}
	\caption{Informal description of the 2nd-order method
		described in Sec.~\ref{sec:short:optim}. \\
		Let $u_t(\cdot)$ be a function computing a direction of descent $\bu_t$
		from a gradient $\bg_t$
		and $\bU_t = \mathrm{Diag}(\bu_t)$.} \label{alg:main_reduc}
	\begin{algorithmic}
		\State{Hyperparameters: $\lambda, \lambda_{\mathrm{int}}$}
		\State{$\mathcal{D}_{\mathrm{g}}, \mathcal{D}_{\mathrm{newt}}:$ independent samplers of minibatches}
		\For{$t$ $\in [1, T]$}
		\State{$Z_t \sim \mathcal{D}_{\mathrm{g}}, \tilde{Z}_t \sim \mathcal{D}_{\mathrm{newt}}$
			\hfill (sample minibatches)}
		\State{$\bg_t \leftarrow \frac{\dd \caL}{\dd \btheta}(\btheta_t, Z_t)$
			\hfill (backward pass)}
		\State{$\bu_t \leftarrow u_t(\bg_t)$ \hfill (custom direction of descent)}
		\State{$\bar{\bg}_t \leftarrow \mathbf{D}_{\btheta_t}^{(1)}(\bu_t) = \bI_{S:P} \bU_t \frac{\dd \mathcal{L}}{\dd \btheta}
			(\btheta_{t}, \tilde{Z}_t)$}
		\State{$\bar{\bH}_t \leftarrow \mathbf{D}_{\btheta_t}^{(2)}(\bu_t) = \bI_{S:P} \bU_t \frac{\dd^2 \mathcal{L}}{\dd \btheta^2}
			(\btheta_{t}, \tilde{Z}_t) \bU_t \bI_{P:S}$}
		\State{$\bD_t \leftarrow \mathrm{Diag}(|\mathbf{D}_{\btheta_t}^{(3)}(\bu_t)|_{iii}^{1/3} :
			i \in \{1, \cdots , S\}) \in \mathbb{R}^{S^2}$}
		\State{$\bfeta_t \leftarrow$ sol.\ of $\bfeta =
			\left(\bar{\bH}_t + \frac{\lambda_{\mathrm{int}}}{2} \|\bD_t \bfeta\| \bD_t^2\right)^{-1}\bar{\bg}_t$
		with max.\ norm $\|\bD_t \bfeta\|$
			\hfill (Method \ref{meth:order2})}
		\State{$\btheta_{t + 1} \leftarrow \btheta_t 
			- \lambda \bU_t \bI_{P:S} \bfeta_t$
			\hfill (training step)}
		\EndFor
	\end{algorithmic}
\end{algorithm}

\paragraph{Setup.}
We consider 4 image classification setups: 
\begin{itemize}
	\item \textbf{MLP}: multilayer perceptron trained on MNIST with
	layers of sizes 1024, 200, 100, 10, and $\tanh$ activation;
	\item \textbf{LeNet}: LeNet-5 \cite{lecun1998gradient} model trained
	on CIFAR-10 with 2 convolutional layers of sizes 6, 16, and 
	3 fully connected layers of sizes 120, 84, 10;
	\item \textbf{VGG}: VGG-11' trained on CIFAR-10.
	VGG-11' is a variant of VGG-11 \cite{simonyan2014very}
	with only one fully-connected layer at the end, instead of 3, 
	with $\mathrm{ELU}$ activation function \cite{clevert2015fast}, without batch-norm;
	\item \textbf{BigMLP}: multilayer perceptron trained on CIFAR-10,
	with 20 layers of size 1024 and one classification layer of size 10, 
	with $\mathrm{ELU}$ activation function.
\end{itemize}
And we have tested 3 optimization methods:
\begin{itemize}
	\item \textbf{Adam}: learning rate selected by grid-search;
	\item \textbf{K-FAC}: learning rate and damping selected by grid-search;
	\item \textbf{NewtonSummary} (ours): $\lambda_1$ and 
	$\lambda_{\mathrm{int}}$ selected by grid search.
\end{itemize}

\paragraph{Results.}
The evolution of the training loss is plotted in Figure \ref{fig:training_curves} 
for each of the 3 optimization methods, for 5 different seeds.
In each set of experiments, the training is successful, 
but slow or unstable at some points.
Anyway, the minimum training loss achieved by Method \ref{meth:order2} (NewtonSummary)
is comparable to the minimum training loss achieved by K-KAC or Adam
in all the series except for MLP, whose training is slower.
We provide the results on the test set in Appendix \ref{app:test}
and a comparison of the training times in Appendix \ref{app:times}.

Some runs have encountered instabilities due to very large step sizes $\bfeta_*$. 
In fact, we did not
use any safeguards, such as a regularization term $\lambda \bI$ added to $\bar{\bH}$,
or clipping the learning rates
to avoid increasing the number of hyperparameters.

\begin{figure}[ht]
	\centering
	\begin{subfigure}{.48\linewidth}
		\includegraphics[width=\linewidth]{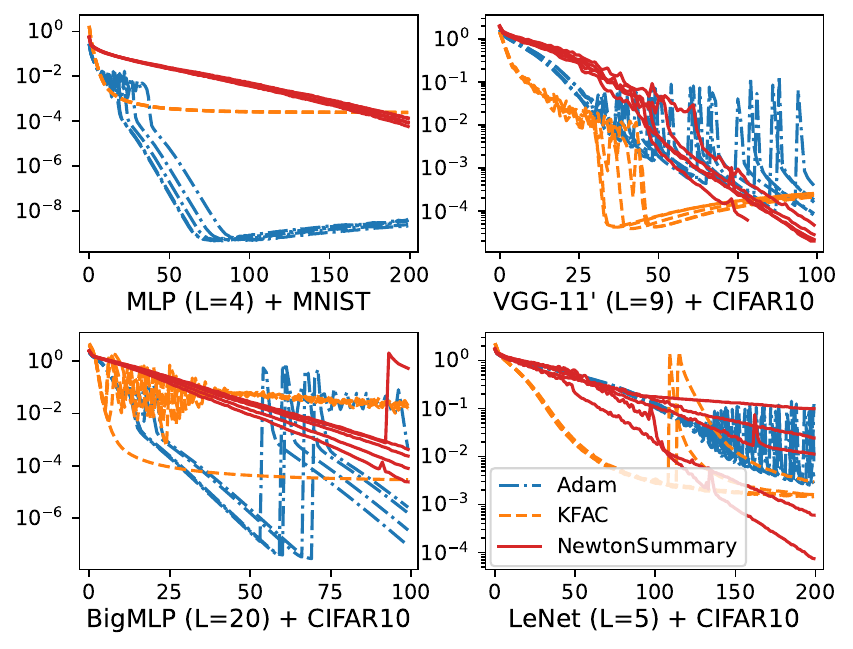}
		\subcaption{Training curves in different setups.
			The reported loss is the negative log-likelihood computed on the training set.}
		\label{fig:training_curves}
	\end{subfigure}
	$\;$
	\begin{subfigure}{.48\linewidth}
		\includegraphics[width=\linewidth]{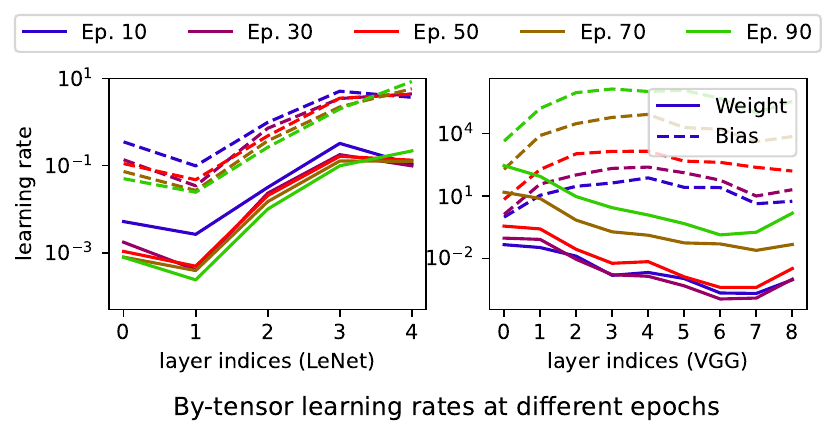}
		\subcaption{Setup: LeNet, VGG-11' trained by SGD on CIFAR-10. 
			Learning rates $\eta_*$ computed according to \eqref{eqn:nesterov},
			specific to each tensor of weights and tensor of biases of each layer.
			For each epoch $k \in \{10, 30, 50, 70, 90\}$, the reported value
			has been averaged over the epochs $\lbrack k - 10, k + 9 \rbrack$ to remove the noise.}
		\label{fig:lrs}
	\end{subfigure}
	\caption{Training curves: Method \ref{meth:order2} (solid lines) versus its diagonal approximation
		(dotted lines) with various hyperparameters.}
	\label{fig:training}
\end{figure}

\paragraph{Extension to very large models.}
Since the matrix $\bar{\bH}$ can be computed numerically
as long as $S$ remains relatively small,
this method may become unpractical for very large models.
However, Method \ref{meth:order2} is flexible enough to
be adapted to such models: one can regroup tensors ``of the same
kind'' to build a coarser partition of the parameters, 
and thus obtain a small $S$, which is exactly what is needed to compute 
$\bar{\bH}$ and invert it.
The difficulty would then be to find a good partition of the parameters,
by grouping all the tensors that ``look alike''. 
We provide an example in Appendix \ref{app:vbig}
with a very deep multilayer perceptron.

\paragraph{Choice of the partition.}
We propose in Appendix \ref{app:partition} an empirical study and a discussion
about the choice of the partition of the parameters. We show
how it affects the training time and the final loss.

\paragraph{Importance of the interactions between layers.}
We show in Appendix \ref{app:diagonal} that the interactions between layers cannot 
be neglected when using our method:
Method \ref{meth:order2} outperforms its \emph{diagonal approximation} on LeNet and VGG11', 
showing the importance of off-diagonal coefficients of $\bar{\bH}$.

\section{Discussion} \label{sec:discussion}

We have shown that it is possible to obtain 2nd and 3rd order information about the loss, 
and that this information can be used to construct per-layer learning rates 
and an optimization method with interesting properties. 
However, this optimization method can only be seen as a proof of concept, 
showing that higher-order derivatives are accessible and can be used to train neural networks, 
and not as a generic optimizer with excellent results on a wide range of tasks and models. 
Therefore, we propose future research directions.

\paragraph{Convergence rate.}
Method \ref{meth:order2} does not come with a precise convergence rate.
The rate proposed in Appendix \ref{app:rate} (Theorem \ref{thm:rate}) gives only a heuristic.
Given the convergence rates of Newton's method and Cauchy's steepest
descent, we can expect to find some in-between convergence rates. 
Since Cauchy's steepest method is vulnerable
to a highly anisotropic Hessian, it would be valuable to know how much
this weakness is overcome by our method.

\paragraph{Accounting for the noise during training.}
Our optimization method remains subject to instabilities during training, which is 
expected for a second-order method, but not acceptable for the end user. 
In fact, it is very likely that our algorithm would achieve better performance 
if it were designed from the beginning to work in a stochastic context. 
Currently, it is designed as if training was done in full batch.

\begin{credits}
	\subsubsection{\ackname} The project leading to this work has received funding from 
	the French National Research Agency (ANR-21-JSTM-0001 and ANR-19-CHIA-0021).
	This work was granted access to the HPC resources of IDRIS under the allocation
	2024-AD011013762R2 made by GENCI.
	We thank Julyan Arbel, Michael N.\ Arbel, Gilles Blanchard and Christophe Giraud
	for their support.
	
	\subsubsection{\discintname}
	The authors have no competing interests to declare that are
	relevant to the content of this article. 
\end{credits}

\bibliographystyle{splncs04}
\bibliography{GroupedNewton}

\newpage

\appendix

\section{Extensions of Pearlmutter's trick} \label{app:fast_comp}

In this appendix, we show how to use the trick of \cite{pearlmutter1994fast}
to compute the terms of the Taylor expansion of $\caL$ and the 
tensor $\mathbf{D}^{d}_{\btheta}(\bu)$ defined in Eqn.~\eqref{eqn:defD}.

\paragraph{Fast computation of the terms of the Taylor expansion.} We recall the Taylor expansion:
\begin{align}
	\caL(\btheta + \bu) = \caL(\btheta) + \sum_{d = 1}^{D} \frac{1}{d!}
	\frac{\dd^d \caL}{\dd \btheta^d}(\btheta)[\bu, \cdots , \bu] 
	+ o(\|\bu\|^D) .
\end{align}
We want to compute:
\begin{align}
	\tilde{\mathbf{D}}_{\btheta}^{d}(\bu) 
	:=\frac{\dd^d \caL}{\dd \btheta^d}(\btheta)[\bu, \cdots , \bu]\in \mathbb{R} .
\end{align}
To do this, we will use:
\begin{align}
	\tilde{\mathbf{D}}_{\btheta}^{d-1}(\bu) = \frac{\dd^{d-1} \caL}{\dd \btheta^{d-1}}(\btheta)
	[\bu, \cdots , \bu]\in \mathbb{R} .
\end{align}
We use the following recursion formula:
\begin{align}
	\tilde{\mathbf{D}}_{\btheta}^{d}(\bu)
	= \left(\frac{\dd \tilde{\mathbf{D}}_{\btheta}^{d-1}(\bu)}{\dd \btheta} \right)^T \bu .
\end{align}
Therefore, at each step $d$, we only have to compute the gradient of 
a scalar $\tilde{\mathbf{D}}_{\btheta}^{d}(\bu)$ according to $\btheta \in \mathbb{R}^P$,
and compute a dot product in the space $\mathbb{R}^P$.
So, computing $\tilde{\mathbf{D}}_{\btheta}^{d}(\bu)$
has a complexity proportional to $d \times P$,
and does not require the computation of the full 
tensor $\frac{\dd^d \caL}{\dd \btheta^d}(\btheta) \in \mathbb{R}^{P^d}$.

\paragraph{Fast computation of $\mathbf{D}^{d}_{\btheta}(\bu)$.} 
We assume that the parameter $\btheta$ is represented
by a sequence of vectors $(\bT^1, \cdots, \bT^S)$:
each coordinate $\theta_i$ belongs to exactly one of the $\bT^s$.
In the same way, given a direction $\bu \in \mathbb{R}^P$ in the space of the parameters,
$\bu$ can be represented by a sequence of vectors $(\bU^1, \cdots, \bU^S)$.

We want to compute the tensor $\mathbf{D}^{d}_{\btheta}(\bu)
\in \mathbb{R}^{S^d}$, whose coefficients are:
\begin{align}
	(\mathbf{D}_{\btheta}^{d}(\bu))_{s_1, \cdots, s_d}
	&= \frac{\partial^d \caL}{\partial \bT^{s_1} \cdots \partial \bT^{s_d}}(\btheta)
	[\bU^{s_1}, \cdots , \bU^{s_d}] ,
\end{align}
for each multi-index $(s_1, \cdots, s_d) \in \{1, \cdots , S\}^d$.

Let us assume that $\mathbf{D}^{d-1}_{\btheta}(\bu)$ is available. 
We can compute the coefficients of $\mathbf{D}_{\btheta}^{d}(\bu)$
as follows:
\begin{align}
	(\mathbf{D}_{\btheta}^{d}(\bu))_{s_1, \cdots, s_d}
	= \left(\frac{\partial (\mathbf{D}^{d-1}_{\btheta}(\bu))_{s_1, \cdots, s_{d-1}}}{
		\partial \bT^{s_d}}\right)^T \bU^{s_d}
\end{align} 
That way, the tensor $\mathbf{D}_{\btheta}^{d}(\bu)$ can be computed
without having to compute the full derivative $\frac{\dd^d \mathcal{L}}{\dd \btheta^d}$.
In fact, we do not need to store objects of size greater than $S^{d-1} \times P$: 
the last operation requires storing $\mathbf{D}^{d-1}_{\btheta}(\bu)$,
which is of size $S^{d-1}$, and the gradient of
each of its elements with respect to $(\bT^1, \cdots, \bT^S)$, which is 
of size $P$.

\section{Derivation of the optimal learning rates} \label{app:derivation}

We consider an update of $\btheta$ with one learning rate $\eta_s$
for each subset $\mathcal{I}_s$ of parameters.
Let $\bI_{S:P} \in \mathbb{R}^{S \times P}$ be the \emph{partition matrix},
verifying $(\bI_{S:P})_{sp} = 1$ if $p \in \mathcal{I}_s$ and
$0$ otherwise, and $\bI_{P:S} := \bI_{S:P}^T$.
We consider an update based on a given direction $\bu_t$ and
we define $\bU_t := \mathrm{Diag}(\bu_t)$:
\begin{align}
	\btheta_{t + 1} = \btheta_t - \bU_t \bI_{P:S} \bfeta ,
\end{align}
where $\bfeta = (\eta_1, \cdots , \eta_S) \in \mathbb{R}^S$.

The second-order approximation of $\caL$ gives:
\begin{align*}
	\caL(\btheta_{t + 1}) &= \caL(\btheta_t - \bU_t \bI_{P:S} \bfeta) \\
	&= \caL(\btheta_t) - \bfeta^T \bI_{S:P} \bU_t \frac{\dd \caL}{\dd \btheta}(\btheta_t) \\
	&\hphantom{=}\; + \frac{1}{2} \bfeta^T \bI_{S:P} \bU_t
	\frac{\dd^2 \caL}{\dd \btheta^2}(\btheta_t) \bU_t \bI_{P:S} \bfeta + o(\|\bfeta\|^2) \\
	&= \caL(\btheta_t) - \bfeta^T \bI_{S:P} \bU_t \bg_t \\
	&\hphantom{=}\; + \frac{1}{2} \bfeta^T \bI_{S:P} \bU_t \bH_t \bU_t \bI_{P:S} \bfeta + 
	o(\|\bfeta\|^2) \\
	&= \caL(\btheta_t) - \bfeta^T \bar{\bg}_t
	+ \frac{1}{2} \bfeta^T \bar{\bH}_t \bfeta + o(\|\bfeta\|^2),
\end{align*}
where: 
\begin{align}
	\bar{\bg}_t := \bI_{S:P} \bU_t \bg_t  \in \mathbb{R}^S , \qquad 
	\bar{\bH}_t := \bI_{S:P} \bU_t \bH_t \bU_t \bI_{P:S} 
	\in \mathbb{R}^{S \times S}.
\end{align}

Now, we omit the $o(\|\bfeta\|^2)$ term and we want to minimize 
according to $\bfeta$ the variation of the loss:
\begin{align}
	\caL(\btheta_{t + 1}) - \caL(\btheta_t) \approx 
	\bDelta_2(\bfeta) := \caL(\btheta_t) - \bfeta^T \bar{\bg}_t
	+ \frac{1}{2} \bfeta^T \bar{\bH}_t \bfeta .
\end{align}
We have:
$\frac{\dd \bDelta_2}{\dd \bfeta} = - \bar{\bg}_t
+ \bar{\bH}_t \bfeta$,
which is zero if, and only if:
$\bar{\bg}_t = \bar{\bH}_t \bfeta$.
If this linear system can be inverted, one can choose:
\begin{align}
	\bfeta = \bfeta_t^* := \bar{\bH}_t^{-1} \bar{\bg}_t . \label{eqn:app:step}
\end{align}

\paragraph{Interpretation as multivariate optimization.}
This method can also be derived by taking the point of view of
multivariate optimization. 
Within our setup, $\btheta$ is considered as a tuple of 
tensors $(\bT^1, \cdots, \bT^S)$. Thus, we want to
minimize the loss $\caL$ with respect to $(\bT^1, \cdots, \bT^S)$.
By abuse of notation, we will write:
\begin{align}
	\caL(\btheta) = \caL(\bT^1, \cdots, \bT^S) .
\end{align}

Now, we assume that we dispose of a direction of descent $-\bU^s$ for
each $\bT^s$. Thus, we can consider the following training step:
\begin{align}
	\forall s \in \{1, \cdots S\}, \quad 
	\bT^s \leftarrow \bT^s - \eta_s \bU^s,
\end{align}
where $(\eta_1, \cdots \eta_S)$ are learning rates.
Thus, the loss change after the training step is:
\begin{align}
	f(\eta_1, \cdots, \eta_S) := \caL(\bT^1 - \eta_1 \bU^1, \cdots, \bT^S - \eta_S \bU^S) - \caL(\bT^1, \cdots, \bT^S) .
\end{align}
When we do a second-order approximation, the loss change becomes:
\begin{align}
	f(\eta_1, \cdots, \eta_S) &\approx - \sum_{s = 1}^{S} \eta_s \left(\frac{\partial \caL}{\partial \bT^s}\right)^T
	\bU^s + \frac{1}{2} \sum_{s_1 = 1}^{S} \sum_{s_2 = 1}^{S}
	\eta_{s_1} \eta_{s_2} (\bU^{s_1})^T \frac{\partial^2 \caL}{\partial \bT^{s_1}\partial \bT^{s_2}}\bU^{s_2} \\
	&\approx  - \bfeta^T \bar{\bg}
	+ \frac{1}{2} \bfeta^T \bar{\bH} \bfeta , \label{eqn:app:alt1}
\end{align}
where $\bfeta = (\eta_1, \cdots , \eta_S) \in \mathbb{R}^S$, $\bar{\bg} \in \mathbb{R}^S$ is the gradient of $f$
and $\bar{\bH} \in \mathbb{R}^{S \times S}$ is the Hessian of $f$:
\begin{align}
	\bar{g}_s &= \frac{\partial f}{\partial \eta_s} = \left(\frac{\partial \caL}{\partial \bT^s}\right)^T \bU^s &
	\bar{H}_{s_1 s_2} &= \frac{\partial^2 f}{\partial \eta_{s_1} \partial \eta_{s_2}}
	= (\bU^{s_1})^T \frac{\partial^2 \caL}{\partial \bT^{s_1}\partial \bT^{s_2}}\bU^{s_2} .
\end{align}
Finally, one can minimize the order-2 approximation of $f$ (Eqn.~\eqref{eqn:app:alt1}) with respect to $\bfeta$,
with any numerical or analytical technique. If done analytically, we roll back to Eqn.~\eqref{eqn:app:step}.

\paragraph{Interpretation as optimization constrained to a vector subspace.}
In addition, the training step can be interpreted as an optimization 
of the descent direction within a vector subspace.
We assume that we dispose of a direction of descent $-\bu \in \mathbb{R}^P$.
Let $(\bu_1, \cdots, \bu_S)$ be a family of vectors of $\mathbb{R}^P$ 
defined by:
\begin{align}
	\forall s, \quad \bu_s = 
		(0_{P_1}, \cdots, 0_{P_{s-1}}, \bU^s, 0_{P_{s+1}}, \cdots, 0_{P_S}) ,
\end{align}
where $P_i$ is the size of the tensor $\bU^i$ (or of $\bT^i$)
and $0_{P_i}$ is the null tensor of size $P_i$.

We want to minimize the loss decrease after one training step 
with respect to the step of descent $\bv$, which is assumed to be small:
\begin{align}
	g(\bv) = \caL(\btheta - \bv) - \caL(\btheta) ,
\end{align}
under the condition $\bv \in \mathrm{span}(\bu_1, \cdots, \bu_S)$.
An order-2 approximation of $g$ gives:
\begin{align*}
	g(\bv) \approx \tilde{g}(\bv) := -\bv^T \frac{\dd \caL}{\dd \btheta}(\btheta) + \frac{1}{2}
	\bv^T \frac{\dd^2 \caL}{\dd \btheta^2}(\btheta) \bv .
\end{align*}
Now, we look for the vector $\bv^*$ such that:
\begin{align}
	\bv^* = \argmin_{\bv^* \in \mathrm{span}(\bu_1, \cdots , \bu_S) }
	\left(-\bv^T \frac{\dd \caL}{\dd \btheta}(\btheta) + \frac{1}{2}
	\bv^T \frac{\dd^2 \caL}{\dd \btheta^2}(\btheta) \bv \right) .
\end{align}
The solution is given by:
\begin{align}
	\bv^* := \bU \bI_{P:S} \bar{\bH}^{-1} \bar{\bg} = \bU \bI_{P:S} \bfeta^* ,
\end{align}
where $\bU = \mathrm{Diag}(\bu)$. So, we recover the direction computed in
Section \ref{sec:optim:presentation}.

\section{Link with Cauchy's steepest descent and Newton's method} \label{app:link_cauchy}

\paragraph{Cauchy's steepest descent.}
Let us consider the trivial partition: $S = 1$, $\mathcal{I}_1 = \{1, \cdots , P\}$.
So, $\bI_{S:P} = (1, \cdots, 1) = \mathds{1}_S^T$. Therefore, the training step is:
\begin{align}
	\btheta_{t + 1} := \btheta_t - \bG_t \mathds{1}_S
	(\mathds{1}_S^T \bG_t \bH_t \bG_t \mathds{1}_S)^{-1} \mathds{1}_S^T \bG_t \bg_t 
	= \btheta_t - \bg_t \frac{\bg_t^T \bg_t}{\bg_t^T \bH_t \bg_t} ,
\end{align}
since $\bG_t \mathds{1}_S = \bg_t$.
We recover Cauchy's steepest descent.

\paragraph{Newton's method.}
Since we aim to recover Newton's method, we assume 
that the Hessian $\bH_t$ is positive definite.
Let us consider the discrete partition: $S = P$, $\mathcal{I}_s = \{s\}$.
So, $\bI_{S:P} = \bI_P$, the identity matrix of $\mathbb{R}^{P \times P}$. 
Therefore, the training step is:
\begin{align}
	\btheta_{t + 1} &:= \btheta_t - \bG_t 
	(\bG_t \bH_t \bG_t)^{-1} \bG_t \bg_t .
\end{align}
To perform the training step, we have to find $\mathbf{x} \in \mathbb{R}^P$
such that: $(\bG_t \bH_t \bG_t)^{-1} \bG_t \bg_t = \mathbf{x}$.
That is, solve the linear system
$\bG_t \bH_t \bG_t \mathbf{x} = \bG_t \bg_t$.
In the case where all the coordinates of the gradient $\bg_t$ are nonzero, 
we can write:
\begin{align}
\mathbf{x} = \bG_t^{-1} \bH_t^{-1} \bG_t^{-1}\bG_t \bg_t = \bG_t^{-1} \bH_t^{-1} \bg_t ,
\end{align}
so the training step becomes:
\begin{align}
\btheta_{t + 1} &:= \btheta_t - \bG_t \mathbf{x} = \btheta_t - \bH_t^{-1} \bg_t,
\end{align}
which corresponds to Newton's method.

\section{Anisotropic Nesterov cubic regularization} \label{app:nesterov}

Let $\bD$ be a diagonal matrix whose diagonal coefficients are all
strictly positive: $\bD = \mathrm{Diag}(d_1, \cdots, d_S)$,
with $d_i > 0$ for all $i$.

We want to minimize the function:
\begin{align}
	T(\bfeta) := - \bfeta^T \bar{\bg} + \frac{1}{2} \bfeta \bar{\bH} \bfeta 
	+ \frac{\lambda_{\mathrm{int}}}{6} \|\bD \bfeta\|^3 .
\end{align}
The function $T$ is strictly convex if, and only if,
$\bar{\bH}$ is positive definite.
Moreover, $T$ is differentiable twice and has at least one global minimum $\bfeta_*$,
so $\frac{\dd T}{\dd \bfeta}(\bfeta_*) = 0$.
Therefore, we first look for the solutions of the equation 
$\frac{\dd T}{\dd \bfeta}(\bfeta) = 0$.

We have:
\begin{align*}
	\frac{\dd T}{\dd \bfeta}(\bfeta) &= -\bar{\bg} + \bar{\bH} \bfeta 
	+ \frac{\lambda_{\mathrm{int}}}{2} \|\bD \bfeta\| \bD^2 \bfeta \\
	&= -\bar{\bg} + \left(\bar{\bH} 
	+ \frac{\lambda_{\mathrm{int}}}{2} \|\bD \bfeta\| \bD^2\right) \bfeta ,
\end{align*}
which is equal to zero if, and only if:
\begin{align}
	\bar{\bg} =
	\left(\bar{\bH} + \frac{\lambda_{\mathrm{int}}}{2} \|\bD \bfeta\| \bD^2\right)\bfeta. 
	\label{eqn:nesterov:linear}
\end{align}
Let $\bfeta' := \bD \bfeta$. Eqn.~\eqref{eqn:nesterov:linear} is then equivalent to:
\begin{align*}
	\bar{\bg} &= \left(\bar{\bH} \bD^{-1}
	+ \frac{\lambda_{\mathrm{int}}}{2} \|\bfeta'\| \bD\right) \bfeta' .  \\
	&= \frac{\lambda_{\mathrm{int}}}{2} \bD \left( \frac{2}{\lambda_{\mathrm{int}}} \bD^{-1} \bar{\bH} \bD^{-1}
	+ \|\bfeta'\| \bI \right) \bfeta'
\end{align*}
Let $\bK := \frac{2}{\lambda_{\mathrm{int}}} \bD^{-1} \bar{\bH} \bD^{-1}$.
We want to solve:
\begin{align}
	\bar{\bg} &= \frac{\lambda_{\mathrm{int}}}{2} \bD \left( \bK + \|\bfeta'\| \bI \right) \bfeta'
	\label{eqn:nesterov:simple_3}
\end{align}
Since $\bK$ is positive definite if, and only if, $\bar{\bH}$ is positive definite,
we consider the following cases.

\paragraph{Case 1: $\bar{\bH}$ is positive definite.}
In this case, Eqn.~\eqref{eqn:nesterov:simple_3} is equivalent to:
\begin{align*}
\bfeta' &= \frac{2}{\lambda_{\mathrm{int}}} \left( \bK + \|\bfeta'\| \bI \right)^{-1} \bD^{-1} \bar{\bg} .
\end{align*}
Now, let $r = \|\bfeta'\|$. We want to solve:
\begin{align}
r = \frac{2}{\lambda_{\mathrm{int}}} 
\left\|\left( \bK + r \bI \right)^{-1} \bD^{-1} \bar{\bg} \right\| . \label{eqn:nesterov:simple_4}
\end{align}
Trivially: $\bfeta$ solution of \eqref{eqn:nesterov:linear}
$\Rightarrow$ $\bD \bfeta$ solution of \eqref{eqn:nesterov:simple_3}
$\Rightarrow$ $\|\bD \bfeta\|$ solution of \eqref{eqn:nesterov:simple_4}.
Reciprocally: $r$ solution of \eqref{eqn:nesterov:simple_4}
$\Rightarrow$ $\bfeta' := (\bar{\bH} \bD^{-1}
+ \frac{\lambda_{\mathrm{int}}}{2} r \bD)^{-1} \bar{\bg}$ solution of \eqref{eqn:nesterov:simple_3}
$\Rightarrow$ $\bD^{-1} \bfeta'$ solution of \eqref{eqn:nesterov:linear}.

Therefore, in order to find the unique global minimum of $T$,
it is sufficient to solve Eqn.~\eqref{eqn:nesterov:simple_4}. 
This is doable numerically.

\paragraph{Case 2: $\bar{\bH}$ is not positive definite.}

We follow the procedure proposed in \cite[Section 5]{nesterov2006cubic}.
Let $\lambda_{\mathrm{min}}$ be the minimum eigenvalue of $\bK$.
So, $\lambda_{\mathrm{min}} \leq 0$. 
Following \cite{nesterov2006cubic}, we look for the unique $\bfeta'$ 
belonging to $\mathcal{C} := 
\{\bfeta' \in \mathbb{R}^S : \|\bfeta'\| > |\lambda_{\mathrm{min}}|\}$,
which is also the solution of maximum norm of Eqn.~\eqref{eqn:nesterov:simple_3}.
Conditionally to $\bfeta' \in \mathcal{C}$, 
$(\bK + \|\bfeta'\| \bI )$ is invertible. So
we only need to solve:
\begin{align}
	r > |\lambda_{\mathrm{min}}| : \quad 
	r = \frac{2}{\lambda_{\mathrm{int}}} 
	\left\|\left( \bK + r \bI \right)^{-1} \bD^{-1} \bar{\bg} \right\| ,
\end{align}
which has exactly one solution $r_*$. Then, 
we compute $\bfeta_* := \bD^{-1} (\bar{\bH} \bD^{-1}
+ \frac{\lambda_{\mathrm{int}}}{2} r_* \bD)^{-1} \bar{\bg}$.

\section{Invariance by subset-wise affine reparameterization} \label{app:invariance}

\subsection{Motivation} \label{app:ntk}

The choice of the best per-layer parameterization is still a debated question.
On the theoretical side, the standard parameterization cannot be used to train very wide networks,
because it leads to a diverging first gradient step \cite{yang2021tensor}.
Besides, the NTK parameterization is widely used in theoretical works in order to 
manage the infinite-width limit \cite{jacot2018neural,du2019gradient,arora2019fine,lee2019wide,mei2022generalization}.
On the practical side, the standard parameterization is preferred over the NTK one
because it leads to better results, both in terms of training and generalization.

So, there is no consensus about the best layer-wise parameterization. 
Thus, ensuring that a method is invariant by layer-wise affine reparameterization guarantees
that its behavior remains the same whatever the choice of the user (standard or NTK parameterization).

\subsection{Claim}

We consider a parameter $\tilde{\btheta}$ such that 
$\btheta = \varphi(\tilde{\btheta})$,
where $\varphi$ is an invertible map, affine on each subset of parameters.
Therefore, its Jacobian is: $\mathbf{J} = \mathrm{Diag}(\alpha_1, \cdots, \alpha_p)$,
where, for all $1 \leq s \leq S$ and $1 \leq p_1, p_2 \leq P$, 
we have:
\begin{align}
	p_1, p_2 \in \mathcal{I}_s \Rightarrow \alpha_{p_1} = \alpha_{p_2} =: a_s .
\end{align}
Also, let $\bar{\mathbf{J}} = \mathrm{Diag}(a_1, \cdots, a_S)$.

We want to compare the training trajectory of $\mathcal{L}(\btheta)$
and $\mathcal{L}(\varphi(\tilde{\btheta}))$ when using Method \ref{meth:order2}.
For any quantity $\mathbf{x}$ computed with the parameterization $\btheta$, 
we denote by $\tilde{\mathbf{x}}$ its counterpart computed with the parameterization $\tilde{\btheta}$.

We compute $\tilde{\bfeta}_*$. Equation \eqref{eqn:nesterov} gives:
\begin{align}
	\tilde{\bfeta}_* =
	\left(\tilde{\bar{\bH}} + \frac{\lambda_{\mathrm{int}}}{2} \|\tilde{\bD} \tilde{\bfeta}_*\|
	\tilde{\bD}^2\right)^{-1}\tilde{\bar{\bg}} . \label{eqn:app:reparam}
\end{align}
Besides:
\begin{align*}
	\tilde{\bar{\bH}} := \bI_{S:P} \tilde{\bU} \tilde{\bH} \tilde{\bU} \bI_{P:S}, \qquad 
	\tilde{\bar{\bg}} := \bI_{S:P} \tilde{\bU} \tilde{\bg} .
\end{align*}

To go further, we need to do an assumption about the direction $\bu$.
\begin{assumption} \label{assump:invariance}
	We assume that $\bU_t$ is computed in such a way that $\tilde{\bU}_t = \mathbf{J} \bU_t$
	at every step.
\end{assumption}
This assumption holds typically when $\bu_t$ is the gradient at time step $t$.
It holds also when $\bu_t$ is a linear combination of the past gradients:
\begin{align*}
	\bu_1 := \bg_1, \qquad 
	\bu_{t + 1} := \mu \bu_t + \mu' \bg_{t + 1} ,
\end{align*}
which includes the momentum.

To summarize, we have:
\begin{align*}
	\tilde{\bU} = \mathbf{J} \bU, \quad
	\tilde{\bH} = \mathbf{J} \bH \mathbf{J}, \quad
	\tilde{\bg} = \mathbf{J} \bg ,
\end{align*}
So:
\begin{align*}
	\tilde{\bar{\bH}} &= \tilde{\mathbf{J}}^2 \bI_{S:P} \bU \bH 
	\bU \bI_{P:S} \tilde{\mathbf{J}}^2 = \tilde{\mathbf{J}}^2 \bar{\bH} \tilde{\mathbf{J}}^2, \\
	\tilde{\bar{\bg}} &= \tilde{\mathbf{J}}^2 \bI_{S:P} \bU \bg = 
	 \tilde{\mathbf{J}}^2 \bar{\bg},
\end{align*}
since $\mathbf{J}$ and $\bU$ are diagonal.
And, since $\bD_{ii} = |(\mathbf{D}_{\btheta}^{(3)}(\bu))_{iii}|^{1/3}$,
then $\tilde{\bD}_{ii} = a_i^2 \bD_{ii}$,
thus $\tilde{\bD} = \tilde{\mathbf{J}}^2 \bD$.

Thus, Eqn.~\eqref{eqn:app:reparam} becomes:
\begin{align*}
	\tilde{\bfeta}_* =
	\left(\tilde{\mathbf{J}}^2 \bar{\bH} \tilde{\mathbf{J}}^2
	+ \frac{\lambda_{\mathrm{int}}}{2} \|\tilde{\mathbf{J}}^2 \bD 
	\tilde{\bfeta}_*\|
	\tilde{\mathbf{J}}^4 \bD^2\right)^{-1}\tilde{\mathbf{J}}^2 \bar{\bg} ,
\end{align*}
which can be rewritten (since $\tilde{\mathbf{J}}$ is invertible):
\begin{align*}
	\tilde{\mathbf{J}}^2 \tilde{\bfeta}_* = 
	\left(\bar{\bH}	+ \frac{\lambda_{\mathrm{int}}}{2} \| \bD \tilde{\mathbf{J}}^2
	\tilde{\bfeta}_*\| \bD^2\right)^{-1}\bar{\bg} .
\end{align*}

Therefore, $\tilde{\bfeta}_*$ is a solution of Eqn.~\eqref{eqn:nesterov} 
in the parameterization $\tilde{\btheta}$ if, and only if, 
$\tilde{\mathbf{J}}^2 \tilde{\bfeta}_*$ is a solution in the parameterization
$\btheta$. Moreover, $\|\tilde{\bD} \tilde{\bfeta}_*\| = 
\|\bD \mathbf{J}^2 \tilde{\bfeta}_*\|$,
so $\tilde{\bfeta}_*$ is the solution of maximum norm $\|\tilde{\bD} \tilde{\bfeta}_*\|$
of \eqref{eqn:nesterov} with parameterization $\tilde{\btheta}$ iff 
$\tilde{\mathbf{J}}^2 \tilde{\bfeta}_*$ is a the solution of maximum norm 
$\|\bD \mathbf{J}^2 \tilde{\bfeta}_*\|$ of \eqref{eqn:nesterov} with parameterization $\btheta$.

Thus, $\bfeta_* = \tilde{\mathbf{J}}^2 \tilde{\bfeta}_*$, 
and the update step in parameterization $\tilde{\btheta}$ is:
\begin{align*}
	\tilde{\btheta}_{t + 1} &= \tilde{\btheta}_t - \tilde{\bU}_t \bI_{P:S} \tilde{\bfeta}_* \\
	&=\tilde{\btheta}_t - \tilde{\bU}_t \bI_{P:S} \tilde{\mathbf{J}}^{-2} \bfeta_* ,
\end{align*}
which can be rewritten:
\begin{align}
	\mathbf{J}^{-1} \btheta_{t + 1} &= \mathbf{J}^{-1} \btheta_{t}
	- \bU \mathbf{J} \bI_{P:S} \tilde{\mathbf{J}}^{-2} \bfeta_* ,  \label{app:reparam:prefinal}
\end{align}
since $\varphi$ is an affine function with factor $\mathbf{J}$.
Finally, Eqn.~\ref{app:reparam:prefinal} boils down to:
\begin{align}
	\btheta_{t + 1} &= \btheta_{t}
	- \bU \bI_{P:S} \bfeta_* ,
\end{align}
which is exactly Method \ref{meth:order2} in parameterization $\btheta$.

\section{Convergence rate in a simple case} \label{app:rate}

We study the convergence of the method presented in Section \ref{sec:optim:presentation}
(without anisotropic Nesterov's cubic regularization):
\begin{align}
\btheta_{t + 1} &= \btheta_t - \bU_t \bI_{P:S} \bfeta_t  ,&
\bfeta_t &:= \bar{\bH}_t^{-1} \bar{\bg}_t,
\end{align}
where: 
\begin{align*}
	\bar{\bH}_t &:= \bI_{S:P} \bU_t \bH_t \bU_t \bI_{P:S}, &
	\bar{\bg}_t &:= \bI_{S:P} \bU_t \bg_t, \\
	\bH_t &:= \frac{\dd^2 \caL}{\dd \btheta^2}(\btheta_t), &
	\bg_t &:= \frac{\dd \caL}{\dd \btheta}(\btheta_t) , \\
	\bU_t &:= -\bG_t, &&
\end{align*}
that is, the direction $\bu_t$ is given
by the gradient $\bg_t$.

We study this optimization method in the case where $\caL$ is
a positive quadratic form:
\begin{align}
	\caL(\btheta) := \frac{1}{2} \btheta^T \bH \btheta, \label{eqn}
\end{align}
where $\bH$ is positive definite and block-diagonal: 
$\bH = \mathrm{Diag}(\bH_1, \cdots, \bH_S)$.

We consider a partition $(\mathcal{I}_s)_{1 \leq s \leq S}$ of the parameter space consistent
with the block-diagonal structure of $\bH$. In other words, if the coefficient $H_{pp}$
of $\bH$ lies in the submatrix $\bH_s$, then $p \in \mathcal{I}_s$.

\begin{theorem} \label{thm:rate}
	The method has a linear rate of convergence. For any $\btheta_{t} \neq 0$:
	\begin{align*}
	\frac{\caL(\btheta_{t + 1})}{\caL(\btheta_{t})} \leq \max_s\left(\frac{(A_s - a_s)^2}{(A_s + a_s)^2} \right) ,
	\end{align*}
	where $a_s = \min \mathrm{Sp}(\bH_s)$ and $A_s = \max \mathrm{Sp}(\bH_s)$. 
	Moreover, this rate is optimal, since it is possible to build $\btheta_t$
	such that:
	\begin{align*}
	\frac{\caL(\btheta_{t + 1})}{\caL(\btheta_{t})} = \max_s\left(\frac{(A_s - a_s)^2}{(A_s + a_s)^2} \right) .
	\end{align*}
\end{theorem}

Alternatively:
\begin{align*}
\frac{\caL(\btheta_{t + 1})}{\caL(\btheta_{t})} \leq \max_s\left(\frac{(\gamma_s - 1)^2}{(\gamma_s + 1)^2} \right) ,
\end{align*}
where $\gamma_s = A_s / a_s \geq 1$. 

\begin{remark}
	For a given $\bH$, better convergence rates can be achieved by reducing the $(\gamma_s)_s$, 
	that is, choosing partitions $(\mathcal{I}_s)_s$ such that, for all $s$,
	the eigenvalues $(h_{p})_{p \in \mathcal{I}_s}$ are not too spread out.
\end{remark}

In other words, good partitions are partitions such that indices of eigenvalues
close to each other are grouped inside the same subset $\mathcal{I}_s$.
On the contrary, grouping the parameters regardless of the eigenspectrum of $\bH$
may lead to
poor convergence rates, since eigenvalues far from each other may be grouped together, 
leading to a very large $\gamma_s$.

\begin{remark}
	To achieve good convergence rates, one should have some access to
	the eigenspectrum of the Hessian, in order to group together the indices
	of eigenvalues having the same order of magnitude.
\end{remark}

\subsection{Proof of Theorem \ref{thm:rate}}

\begin{proof}

We have:
\begin{align*}
	\caL(\btheta_{t + 1}) &= \frac{1}{2} \btheta_{t + 1}^T \bH \btheta_{t + 1} \\
	&= \frac{1}{2} (\btheta_t - \bG_t \bI_{P:S} \bfeta_t)^T 
	\bH (\btheta_t - \bG_t \bI_{P:S} \bfeta_t) \\
	&= \caL(\btheta_{t}) - \btheta_{t}^T \bH \bG_t \bI_{P:S} \bfeta_t
	+ \frac{1}{2} \bfeta_t^T \bI_{S:P} \bG_t \bH \bG_t \bI_{P:S} \bfeta_t \\
	&= \caL(\btheta_{t}) - \btheta_{t}^T \bH \bG_t \bI_{P:S} \bfeta_t
	+ \frac{1}{2} \bfeta_t^T \bar{\bH}_t \bfeta_t \\
	&= \caL(\btheta_{t}) - \bg_{t}^T \bG_t \bI_{P:S} \bar{\bH}_t^{-1} \bar{\bg}_t
	+ \frac{1}{2} \bar{\bg}_t \bar{\bH}_t^{-1} \bar{\bg}_t \\
	&= \caL(\btheta_{t}) - \frac{1}{2} \bar{\bg}_t^T \bar{\bH}_t^{-1} \bar{\bg}_t .
\end{align*}

Now, we study $\Delta = - \frac{1}{2} \bar{\bg}^T \bar{\bH}^{-1} \bar{\bg}$. 
We omit the time $t$ for the sake of readability.

We can write $\bg$ as a block vector: 
$\bg = (\bg_1, \cdots, \bg_S)$, where $\bg_s \in \mathbb{R}^{|\mathcal{I}_s|}$
for all $1 \leq s \leq S$. 
Thus, since $\bH$ is block-diagonal:
\begin{align*}
	\bar{\bH} &= \mathrm{Diag}(\bg_s^T \bH_s \bg_s : s \in \{1, \cdots, S\}), \\
	\bar{\bH}^{-1} &= \mathrm{Diag}((\bg_s^T \bH_s \bg_s)^{-1} : s \in \{1, \cdots, S\}) .
\end{align*}
Also, $\bar{\bg}_s = \bg_s^T \bg_s$, then:
\begin{align*}
	\Delta &= - \frac{1}{2} \sum_{s = 1}^S \frac{(\bg_s^T \bg_s)^2}{\bg_s^T \bH_s \bg_s} \\
	&= - \frac{1}{2} \sum_{s = 1}^S \frac{(\bg_s^T \bg_s)^2 (\bg_s^T \bH_s^{-1} \bg_s)}{
		(\bg_s^T \bH_s \bg_s) (\bg_s^T \bH_s^{-1} \bg_s)}.
\end{align*}

By Kantorovich's inequality, we have:
\begin{align*}
	\Delta &\leq - \frac{1}{2} \sum_{s = 1}^S \frac{\bg_s^T \bH_s^{-1} \bg_s}{
		\frac{1}{4}(\frac{a_s}{A_s} + \frac{A_s}{a_s} + 2)} \\
	&\leq - 2 \sum_{s = 1}^S \frac{(\bg_s^T \bH_s^{-1} \bg_s) A_s a_s}{
		(A_s + a_s)^2} .
\end{align*}
Thus:
\begin{align*}
	\Delta &\leq -\min\left(\frac{2 A_s a_s}{(A_s + a_s)^2}\right) 
	\sum_{s = 1}^S \bg_s^T \bH_s^{-1} \bg_s \\
	&\leq - \min\left(\frac{2 A_s a_s}{(A_s + a_s)^2}\right) \btheta^T \bH \btheta .
\end{align*}

Finally, when dividing by $\caL(\btheta_t) = \frac{1}{2}\btheta^T \bH \btheta$, we have:
\begin{align*}
	\frac{\caL(\btheta_{t + 1})}{\caL(\btheta_t)} -1 &\leq - \min\left(\frac{4 A_s a_s}{(A_s + a_s)^2}\right) \\
	\frac{\caL(\btheta_{t + 1})}{\caL(\btheta_t)}  &\leq \max\left(\frac{(A_s - a_s)^2}{(A_s + a_s)^2}\right)
\end{align*}

Besides, this rate is optimal, since it is possible to build $\btheta_t$
such that:
\begin{align*}
\frac{\caL(\btheta_{t + 1})}{\caL(\btheta_{t})} = \max_s\left(\frac{(A_s - a_s)^2}{(A_s + a_s)^2} \right) .
\end{align*}

To do so, let $s_0 \in \argmax_s\left(\frac{(A_s - a_s)^2}{(A_s + a_s)^2}\right)$.
Let $\bg_{\min}$ be an eigenvector of $\bH$ associated to $a_{s_0}$
and $\bg_{\max}$ be an eigenvector of $\bH$ associated to $A_{s_0}$, orthogonal
with $\|\bg_{\min}\| = \|\bg_{\max}\| = 1$. 
Also, let $\btheta_{t} = \bH^{-1} (\bg_{\min} + \bg_{\max})$.

Thus:
\begin{align*}
\caL(\btheta_{t + 1}) - \caL(\btheta_{t}) 
= - \frac{1}{2} \frac{(\bg_{s_0}^T \bg_{s_0})^2}{\bg_{s_0}^T \bH_{s_0} 
\bg_{s_0}}
= - \frac{1}{2} \frac{2}{A_{s_0} + a_{s_0}}
\end{align*}
Finally:
\begin{align*}
	\frac{\caL(\btheta_{t + 1})}{\caL(\btheta_{t})} &=
	1 - \frac{1}{2} \frac{2}{A_{s_0} + a_{s_0}} \frac{1}{\frac{1}{2} \bg_{s_0}^T \bH_{s_0}^{-1}
		\bg_{s_0}}\\
	&= 1 - \frac{2}{A_{s_0} + a_{s_0}} \frac{1}{A_{s_0}^{-1} + a_{s_0}^{-1}} \\
	&= 1 - \frac{2 A_{s_0} a_{s_0}}{(A_{s_0} + a_{s_0})^2}  \\
	&= \frac{(A_{s_0} - a_{s_0})^2}{(A_{s_0} + a_{s_0})^2}
\end{align*}

\end{proof}

\section{Experimental details} \label{app:implementation}

\paragraph{Practical implementation.}
To implement the method proposed in Section \ref{sec:short:optim},
we propose Algorithm \ref{alg:main}. The key function are
compute\_lr$(\lambda_{\mathrm{int}} ; 
\bar{\bH}, \bar{\bg}, \bD)$,
which returns a solution $\bfeta_*$ of:
\begin{align*}
\bfeta_* &=
\left(\bar{\bH} + \frac{\lambda_{\mathrm{int}}}{2} \|\bD \bfeta_*\| \bD^2\right)^{-1}\bar{\bg}, \\
\text{with: } \quad \bar{\bH} &:= \bI_{S:P} \mathrm{Diag}(\bu) \frac{\dd^2 \caL}{\dd 
	\btheta^2}(\btheta, \tilde{Z}) \mathrm{Diag}(\bu) \bI_{P:S}, \\
\quad \bar{\bg} &:= \bI_{S:P} \mathrm{Diag}(\bu) \bg_t, \\
\quad \bD &:= \mathrm{Diag}\left(\left(\left|
\bD^{(3)}_{\btheta}(\bu) \right|_{iii}^{1/3}\right)_{1 \leq i \leq S}\right) ,
\end{align*}
and compute\_Hg$(\caL, \btheta, \tilde{Z}, \bu)$,
which returns the current value of $(\bar{\bH}, \bar{\bg}, \bD)$.
``momentum$(\mu, \mathbf{x}, \tilde{\mathbf{x}})$'' 
returns $\mathbf{x}$ if $\tilde{\mathbf{x}}$ is undefined, else
$\mu \tilde{\mathbf{x}} + \mathbf{x}$. ``schedule$(p_{\mathrm{sch}}, f_{\mathrm{sch}};\cdots)$'' 
corresponds to torch.optim.lr\_scheduler.ReduceLROnPlateau
with patience $p_{\mathrm{sch}}$ and factor $f_{\mathrm{sch}}$,
in order to reduce the step size $\lambda_t$ when the loss attains a plateau.%
\footnote{See torch.optim.lr\_scheduler.ReduceLROnPlateau.}
The samplers $\mathcal{D}_g$ and $\mathcal{D}_{\mathrm{newt}}$ are respectively
used to compute the gradients $\bg_t$ and $(\bar{\bH}, \bar{\bg})$ used in ``compute\_lr''.

The hyperparameters are: the initial step size $\lambda_1$, the
momentum $\mu_g$ on the gradients $\bg_t$ (as for the SGD with momentum), 
the minibatch size $B$ to sample the $\tilde{Z}$ (used to compute $\bar{\bg}$, $\bar{\bH}$ and $\bD$), 
the number of steps $\tau$ between each call of compute\_lr,
the averaging window $\upsilon$ for $\bar{\bH}, \bar{\bg}, \bD$
(the average of $\bar{\bH}, \bar{\bg}, \bD$ over the last $\tau \times \upsilon$
steps is used in compute\_lr),
the internal damping $\lambda_{\mathrm{int}}$, and 
the parameters of the scheduler $p_{\mathrm{sch}}$, $f_{\mathrm{sch}}$.

\begin{algorithm}
	\caption{Implementation of the second-order optimization method
		described in Sec.~\ref{sec:short:optim}. \\
		$\lambda_1$ and $\lambda_{\mathrm{int}}$ are the only hyperparameter to be tuned across the experiments, 
		the others are fixed.} \label{alg:main}
	\begin{algorithmic}
		\State{Hyperparams: $\lambda_1, \mu_g, B_g, B, \tau,
			\upsilon, \lambda_{\mathrm{int}}, p_{\mathrm{sch}}, f_{\mathrm{sch}}$}
		\State{$\mathcal{D}_{\mathrm{g}} \leftarrow$ sampler of minibatches of size $B_g$}
		\State{$\mathcal{D}_{\mathrm{newt}} \leftarrow$ sampler of minibatches of size $B$}
		\For{$t$ $\in [1, T]$}
		\State{$Z_t := (X_t, Y_t) \sim \mathcal{D}_{\mathrm{g}}$
			\hfill (sample minibatch)}
		\State{$\caL_t \leftarrow \caL(\btheta_t, Z_t)$ \hfill (forward pass)}
		\State{$\bg_t \leftarrow \frac{\dd \caL}{\dd \btheta}(\btheta_t, Z_t)$
			\hfill (backward pass)}
		\State{$\tilde{\bg}_t \leftarrow$ momentum$(\mu_g; \bg_t, \tilde{\bg}_{t - 1})$}
		\If{$t \, \% \, \tau == 0$}
		\State{sample $\tilde{Z}_t \sim \mathcal{D}_{\mathrm{newt}}$}
		\State{$\bar{\bH}_t, \bar{\bg}_t, \bD_t \leftarrow$ compute\_Hg$(\caL, \btheta_t, \tilde{Z}, \tilde{\bg}_t)$}
		\State{$\bfeta_t \leftarrow$ compute\_lr$(\lambda_{\mathrm{int}} ; 
			\bar{\bH}, \bar{\bg}, \bD)$
			\hfill ($\bar{\bH}, \bar{\bg}, \bD$ averaged on the last $\upsilon$ steps)}
		\State{$\tilde{\bfeta}_t \leftarrow$ momentum$(\mu_{\eta}; (\bfeta_t)_+, 
			\tilde{\bfeta}_{t - 1})$}
		\EndIf
		\State{$\btheta_{t + 1} \leftarrow \btheta_t 
			- \lambda_t \mathrm{Diag}(\tilde{\bg}_t) \bI_{P:S} \tilde{\bfeta}_t$
			\hfill (training step)}
		\State{$\lambda_{t + 1} \leftarrow$ schedule$(p_{\mathrm{sch}}, f_{\mathrm{sch}};
			t, \caL_t, \lambda_t)$}
		\EndFor
	\end{algorithmic}
\end{algorithm}

\paragraph{Explanation.}
The ``momentum'' functions are used to deal with the stochastic part of the training process, since
our method has not been designed to be robust against noise. The period $\tau$
is usually strictly greater than $1$, in order to avoid calling ``compute\_lr'' at every step, 
which would be costly. The minibatch size $B$ should be large enough to reduce noise in the estimation of
$\bfeta_*$. If we denote by $B_{g}$ the size of the minibatches in $\mathcal{D}_{g}$,
then we recommend the following setup: $\tau = \frac{B}{B_{g}}
= \frac{1}{1 - \mu_g}$. That way, we ensure that the training data are sampled from 
$\mathcal{D}_{g}$ and $\mathcal{D}_{\mathrm{newt}}$ at the same rate, and that 
$\tilde{\bg}_t$ memorizes the preceding gradients $\bg_t$ for $\tau$ steps.
Besides, we have to take the positive part $(\bfeta_t)_+$ of $\bfeta_t$ in order to avoid negative learning
rates.

\paragraph{Experimental setup.}
We provide in Table \ref{tbl:hyperparams} the hyperparameters 
fixed for all the experiments. In Table \ref{tbl:hyperparams:all},
we report the results of the grid-search for the hyperparameters
of the 3 tested optimization methods.

\begin{table}[ht]
	\caption{Hyperparameters fixed in all the series of experiments.
		$N_e$ is the number of training steps per epoch.} 
	\label{tbl:hyperparams}
	\vspace*{3mm}
	\centering
	\begin{tabular}{ccccccc}
		\toprule 
		$\mu_g$ & $B_g$ & $B$ & $\tau$ &
		$\upsilon$ & $p_{\mathrm{sch}}$ & 
		$f_{\mathrm{sch}}$ \\
		\midrule 
		$0.9$ & $10^2$ & $10^3$ & $10$ &
		$3$ & $2$ & $0.5$ \\
		\bottomrule
	\end{tabular}
\end{table}

\begin{table}[ht]
	\caption{Hyperparameters tuned for each series of experiments. 
		$\eta$: learning rate, $\lambda_1$: initial step size.} 
	\label{tbl:hyperparams:all}
	\vspace*{3mm}
	\centering
	\begin{tabular}{cccccc}
		\toprule 
		&& MLP & LeNet & VGG-11' & BigMLP \\
		\toprule
		Adam & $\eta$ & $3 \cdot 10^{-4}$ & $3 \cdot 10^{-4}$ & $10^{-5}$ & $10^{-5}$ \\
		\midrule 
		\multirow{2}{*}{KFAC} & $\eta$ & $10^{-4}$ & $10^{-4}$ & $3 \cdot 10^{-4}$ & $10^{-5}$ \\
		& $\lambda$ & $10^{-2}$ & $3 \cdot 10^{-2}$ & $3 \cdot 10^{-2}$ & $10^{-2}$ \\
		\midrule 
		\multirow{2}{*}{Ours} & $\lambda_1$ & $3 \cdot 10^{-2}$ & $3 \cdot 10^{-1}$ & $3 \cdot 10^{-1}$ & $10^{-1}$ \\
		& $\lambda_{\mathrm{int}}$ & $1$ & $1$ & $1$ & $3$ \\
		\bottomrule 
	\end{tabular}
\end{table}

\paragraph{K-FAC update periods.}
In accordance with the K-FAC packages, we have chosen to increase the update period of the pre-conditioner
to reduce the training time.
Specifically, we have chosen to perform a covariance update every $10$ steps,
and the inversion of the Fisher matrix every $100$ steps:
\begin{itemize}
	\item with tensorflow/kfac: use PeriodicInvCovUpdateKfacOpt with: cov\_update\_every = 10 and
	invert\_every = 100;
	\item with alecwangcq/KFAC-Pytorch: use KFACOptimizer with: TCov = 10 and TInv = 100.
\end{itemize}

\FloatBarrier

\section{Very deep multilayer perceptron} \label{app:vbig}

\paragraph{Grouping the layers.}
In addition to the neural networks considered in Section \ref{sec:expes},
we have also tested ``VBigMLP'', a very deep multilayer perceptron with 
100 layers of size 1024 trained on CIFAR-10. 
Instead of considering $S = 2 L = 200$ groups of parameters,
we split the sequence of layers of VBigMLP into 5 chunks.
Then, each chunk is divided into 2 parts, one containing the weight tensors, 
and the other the bias vectors. 
Finally, we have $S = 10$ subsets of parameters, grouped by role (weight/bias)
and by position inside the network.

\paragraph{Experimental results.}
We show in Figure \ref{fig:H_Hinv_vbig} the matrices $\bar{\bH}$
and $\bar{\bH}^{-1}$ at different stages of training.
At initialization, even if the neural network is very deep, 
we observe that all the chunks of the network interact together, 
even the first one with the last one.
However, after several training steps, the long-range interactions seem 
to disappear. Incidentally, the matrices become tridiagonal,
which ties in with the block-tridiagonal approximation of the inverse of the Hessian
done by \cite{martens2015optimizing}.

\begin{figure}[ht]
	\centering
	\begin{subfigure}{.48\linewidth}
		\includegraphics[width=\linewidth]{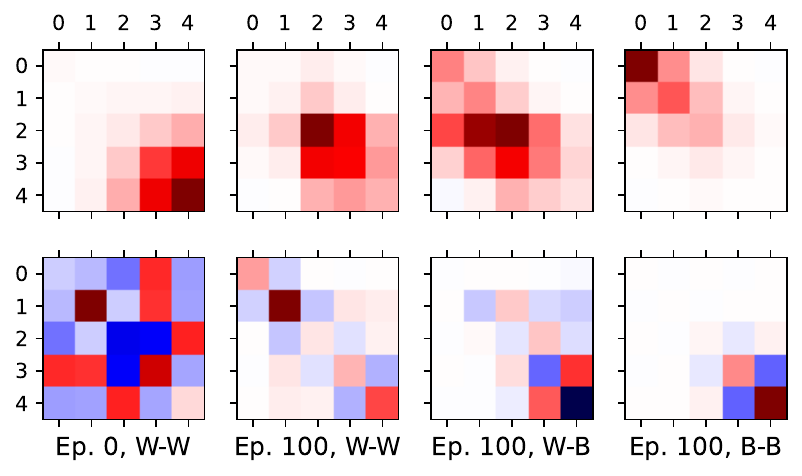}
		\subcaption{Submatrices of $\bar{\bH}$ (first row) and $\bar{\bH}^{-1}$ (second row),
			at initialization and before the 100th epoch.}
		\label{fig:H_Hinv_vbig}
	\end{subfigure}
	$\;$
	\begin{subfigure}{.48\linewidth}
		\includegraphics[width=\linewidth]{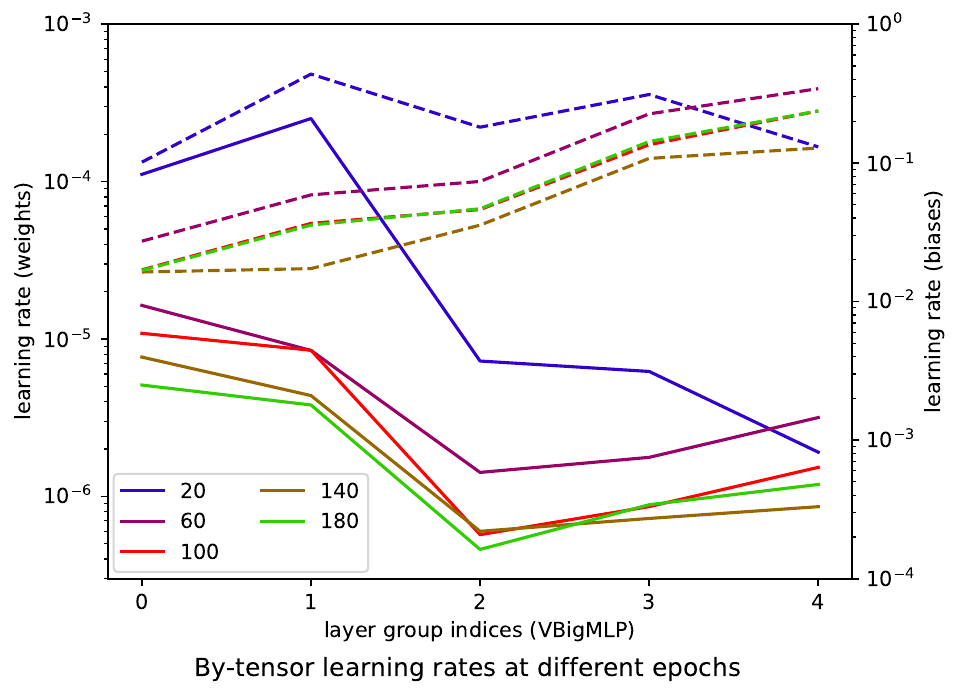}
		\subcaption{
			Learning rates $\eta_*$ computed according to \eqref{eqn:nesterov},
			specific to each subset of parameters.}
		\label{fig:lrs_vbig}
	\end{subfigure}
	\caption{Matrices $\bar{\bH}$ and $\bar{\bH}^{-1}$ and 
		per-subset-of-parameters learning rates obtained with VBigMLP. \\
		Legend for the figure on the right: solid lines: weights; dotted lines: biases. 
		For each epoch $k \in \{20, 60, 100, 140, 180\}$, the reported value
		has been averaged over the epochs $\lbrack k - 20, k + 19 \rbrack$ to remove the noise.}
	\label{fig:vbig_mlp}
\end{figure}

In Figure \ref{fig:lrs_vbig}, we observe the evolution of the 
learning rates $\eta_*$ computed according to \eqref{eqn:nesterov}.
First, there are all decreasing during training.
Second, the biases in the last layers of the network seem to need
larger learning rates than biases in the first layers.
Third, the learning rate computed for the weights of the first 
chunk of layers is smaller than the others.

Finally, the training curves in Figure \ref{fig:training_curves_vbig}
indicate that our method can be used to train very deep networks.
In this setup, it is close to be competitive with Adam.
Besides, we did not manage to tune the learning rate and the damping of K-FAC to 
make it work in this setup.

We have also plotted the evolution of the test loss and test accuracy during training
(see Figure \ref{fig:testing_curves_vbig}).
It is clear that Adam does not generalize at all, while
our method attains a test accuracy around 35 \% -- 40 \%.

\begin{figure}[ht]
	\centering 
	\begin{subfigure}{.48\linewidth}
		\includegraphics[width=\linewidth]{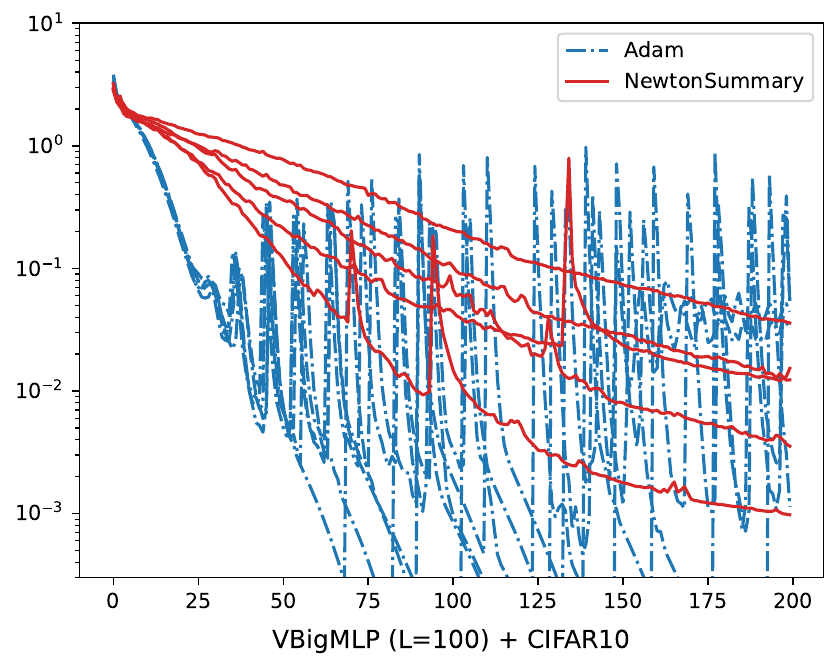}
		\subcaption{Training curves.}
		\label{fig:training_curves_vbig}
	\end{subfigure}
	$\;$
	\begin{subfigure}{.48\linewidth}
		\includegraphics[width=\linewidth]{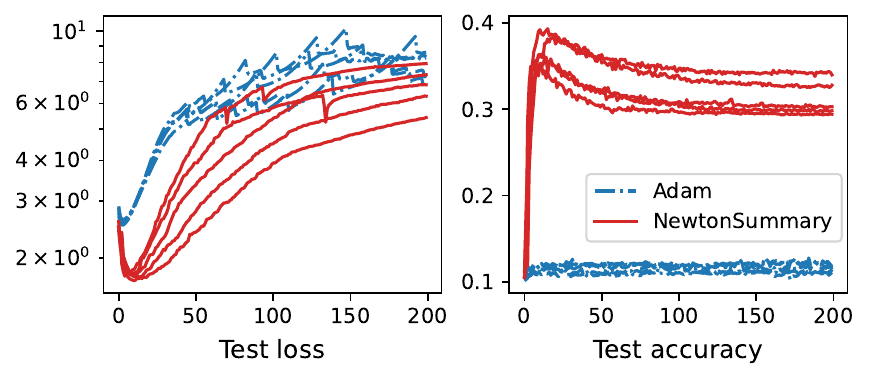}
		\caption{Test loss and test accuracy during training.}
		\label{fig:testing_curves_vbig}
	\end{subfigure}
	\caption{VBigMLP + CIFAR-10.}
	\label{fig:curves_vbig}
\end{figure}

\FloatBarrier

\section{Test loss and test accuracy} \label{app:test}

\begin{figure}[ht]
	\centering 
	\begin{subfigure}{.48\linewidth}
		\includegraphics[width=\linewidth]{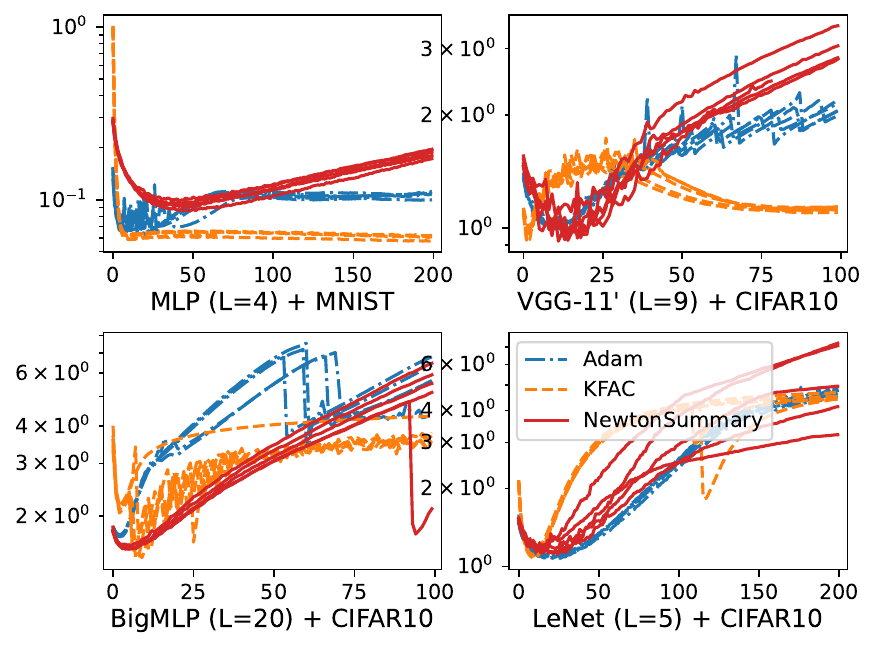}
		\subcaption{Test negative log-likelihood in different setups.}
		\label{fig:testing_curves_nll}
	\end{subfigure}
	$\;$
	\begin{subfigure}{.48\linewidth}
		\includegraphics[width=\linewidth]{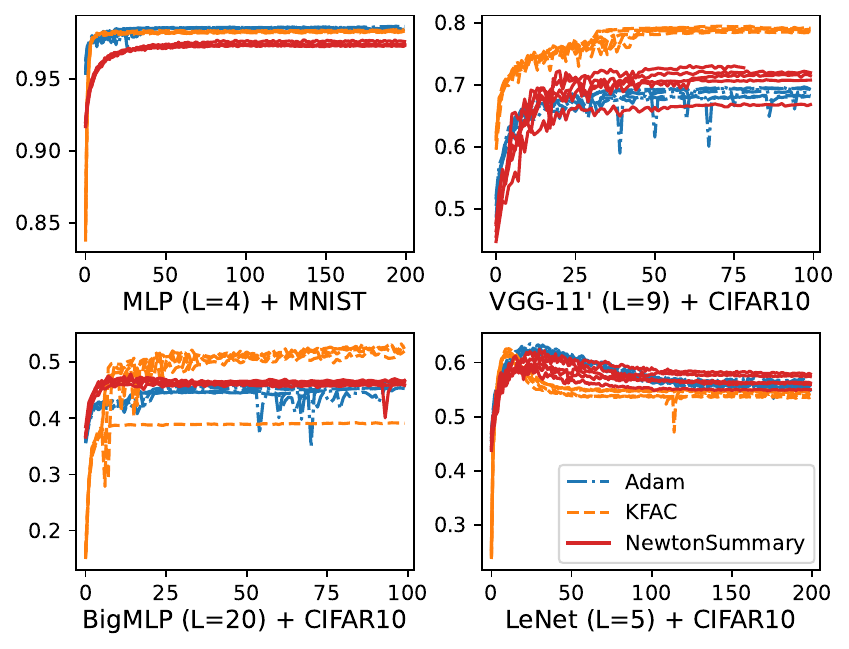}
		\caption{Test accuracy in different setups.}
		\label{fig:testing_curves_acc}
	\end{subfigure}
	\caption{Test metrics in various setups.}
	\label{fig:testing_curves}
\end{figure}

In Figure \ref{fig:testing_curves_nll} and Figure \ref{fig:testing_curves_acc}, we have reported
the test negative log-likelihood and the test accuracy 
of the same experiments as in Section \ref{sec:training} (Figure \ref{fig:training_curves}).

Our method is competitive with Adam and K-FAC when comparing the test losses, except for the MLP trained on MNIST. 
In several cases, we observe a discrepancy between the test loss and the test accuracy: 
one method might be better than another according to the loss, 
but worse in terms of accuracy. 
In particular, the test loss of our method can achieve smaller test losses 
than the other methods, while its test accuracy remains slightly lower (BigMLP, VGG).

\FloatBarrier

\section{Choice of the partition} \label{app:partition}

We have trained VGG-11' on CIFAR-10 using our method
with different partition choices. 
In Table \ref{tbl:partition}, we report the final training losses, 
the training time (wall-time), and the maximum memory usage.

Not surprisingly, the finer the partition, the better the results.
However, this comes at a cost: training with finer partitions takes more time.
We also observe that memory usage tends to decrease as 
the partition becomes finer.

The partitions we have tested are:
\begin{itemize}
	\item trivial, $S = 1$: all the tensors are grouped together;
	\item weights-biases, $S = 2$: all the weights are grouped together,
	and all the biases too;
	\item blocks-$k$, $S = 2 k + 2$: the sequence of convolutional layers
	is split into $k$ consecutive blocks, and each one is split in two (weights + biases); 
	weights and biases of the final fully-connected layer are considered separately 
	(hence the ``$+2$'' in $S$);
	\item alternate-$k$, $S = 2 k + 2$: the convolutional layer $l$
	is put in the $\tilde{s}$-block if $l \% k = \tilde{s}$; then, 
	each block is split in two (weights + biases); 
	weights and biases of the final fully-connected layer are considered separately 
	(hence the ``$+2$'' in $S$);
	\item canonical, $S =\#\text{tensors}$: each tensor is considered separately.
\end{itemize}

\begin{table}[ht]
	\caption{Influence of the choice of the partition
		when training VGG-11' on CIFAR-10.} 
	\label{tbl:partition}
	\vspace*{3mm}
	\centering
	\begin{tabular}{cccc}
		\toprule
		partition & train NLL & time ($\mathrm{s}$) & mem.\ ($\mathrm{Go}$) \\
		\toprule 
		trivial & $8.12 \cdot 10^{-1}$ & $2~512$ & $2.49$ \\
		\midrule 
		weights-biases & $7.64 \cdot 10^{-1}$ & $2~855$ & $2.49$ \\
		\midrule 
		blocks-2 & $5.94 \cdot 10^{-1}$ & $3~182$ & $2.38$ \\
		alternate-2 & $5.70 \cdot 10^{-1}$ & $3~422$ & $2.08$ \\
		\midrule 
		blocks-4 & $1.50 \cdot 10^{-2}$ & $3~674$ & $1.97$ \\
		alternate-4 & $5.37 \cdot 10^{-2}$ & $4~180$ & $1.91$\\
		\midrule 
		canonical & $3.05 \cdot 10^{-4}$ & $4~612$ & $1.88$ \\
		\bottomrule
	\end{tabular}
\end{table}

\FloatBarrier

\section{Measuring the importance of interactions between layers} \label{app:diagonal}

\paragraph{Diagonal approximation of Method \ref{meth:order2}.}
Throughout this paper, we have emphasized the importance of considering 
the interactions between layers when training a neural network.
In fact, Method \ref{meth:order2} allows the user to keep track of them at a 
reasonable computational cost.
But is it useful to take these interactions into account?

If the computational cost is really an issue, one can compute only the diagonal
coefficients of $\bar{\bH}$ and set the off-diagonal coefficients to zero.
Let $\bar{\bH}^0$ be this \emph{diagonal approximation} of $\bar{\bH}$:
\begin{align*}
	\bar{\bH}^0 := \mathrm{Diag}((\bar{h}_{ii})_{1 \leq i \leq S}) ,
\end{align*}
where $(\bar{h}_{ii})_{1 \leq i \leq S}$ are the diagonal coefficients of $\bar{\bH}$.

Then, we call the \emph{diagonal approximation} of Method \ref{meth:order2},
Method \ref{meth:order2} where $\bar{\bH}$ has been replaced by $\bar{\bH}^0$.

\paragraph{Experiments.}
We have tested Method \ref{meth:order2} with the hyperparameters we have used in Section
\ref{sec:training} and its diagonal approximation with a grid of hyperparameters $\lambda_1$
and $\lambda_{\mathrm{int}}$. The results are shown in Figure \ref{fig:training_diagonal}.
Note that the configuration $\lambda_1 = 1$ was tested with VGG11', 
but resulted in instantaneous divergence,
so we have not plotted the corresponding training curves.

According to Figure \ref{fig:training_diagonal}, the diagonal approximation of Method \ref{meth:order2}
performs worse or is more unstable than \ref{meth:order2}.
Therefore, when training LeNet or VGG11' with CIFAR10, it is better to keep the off-diagonal
coefficients of $\bar{\bH}$. 

In short, one should worry about the interactions between layers.

\begin{figure}[ht]
	\centering
	\begin{subfigure}{.48\linewidth}
		\includegraphics[width=\linewidth]{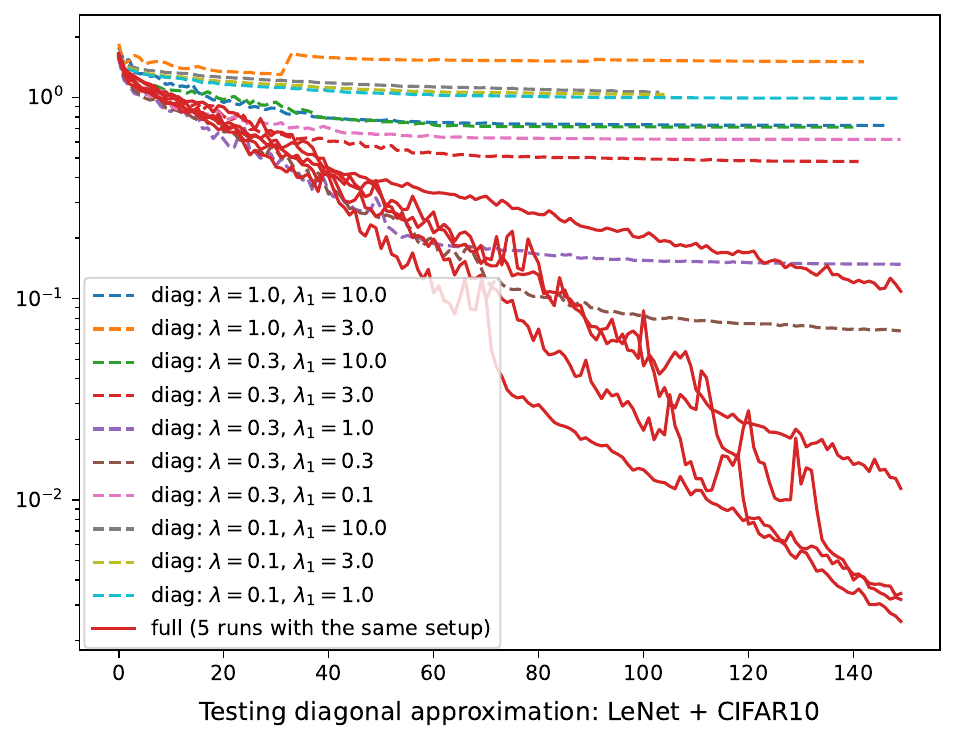}
		\subcaption{LeNet + CIFAR10.}
		\label{fig:training_diagonal_lenet}
	\end{subfigure}
	$\;$
	\begin{subfigure}{.48\linewidth}
		\includegraphics[width=\linewidth]{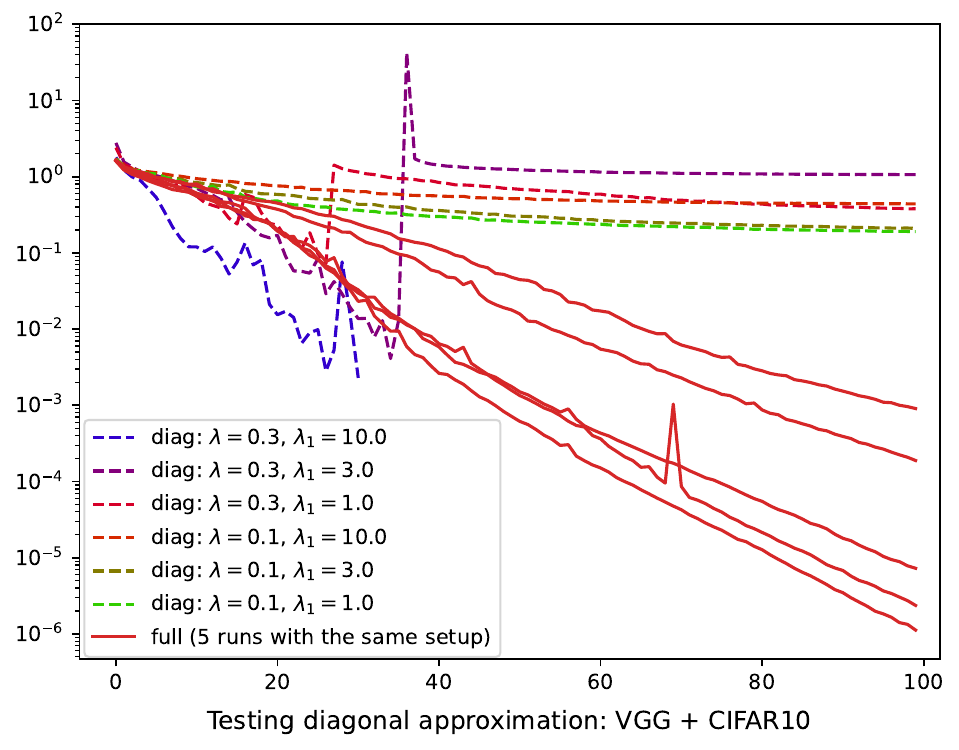}
		\subcaption{VGG11' + CIFAR10.}
		\label{fig:training_diagonal_vgg}
	\end{subfigure}
	\caption{Training curves: Method \ref{meth:order2} (solid lines) versus its diagonal approximation
	(dotted lines) with various hyperparameters.}
	\label{fig:training_diagonal}
\end{figure}
\FloatBarrier

\section{Higher-order derivatives of a multivariate function} \label{app:dieudonne}

In this section, we recall formally the definition of higher-order derivatives of a multivariate function, 
following \cite{dieudonne_foundations_1960}. 

\subsection{Definitions}

Let $\mathcal{L}(E, F)$ be the space of linear maps from $E$ to $F$ and 
$\mathcal{L}_d(E, F)$ be the space of $d$-linear maps from $E \times \cdots \times E$
to $F$. For instance, the space of linear forms on $\mathbb{R}^d$ is
denoted by $\mathcal{L}(\mathbb{R}^P, \mathbb{R})$,
and the space of $3$-linear forms on $\mathbb{R}^P \times \mathbb{R}^P \times \mathbb{R}^P$
is denoted by $\mathcal{L}_3(\mathbb{R}^P, \mathbb{R})$.

Let $f$ be a smooth multivariate function from $\mathbb{R}^P$ to $\mathbb{R}$:
\begin{align}
	f : \mathbb{R}^P \rightarrow \mathbb{R} .
\end{align}

\paragraph{Differential of order 1.}
The differential of $f$ at a point $\btheta \in \mathbb{R}^P$ is the
only linear form $T_f(\btheta) \in \mathcal{L}(\mathbb{R}^P, \mathbb{R})$ such that:
\begin{align}
	\lim_{\bu \rightarrow 0}\frac{f(\btheta + \bu) - f(\btheta) - T_f(\btheta)(\bu)}{\|\bu\|^2} = 0.
\end{align}
Since $T_f(\btheta)$ is a linear form, there exists a vector $\bg \in \mathbb{R}^P$ such that:
\begin{align}
	T_f(\btheta)(\bu) = \bg^T \bu .
\end{align} 
The vector $\bg$ is nothing other than the \emph{gradient} of $f$ at $\btheta$,
and $T_f(\btheta)$ is the \emph{differential} of $f$ at $\btheta$,
that we denote by $\frac{\dd f}{\dd \btheta}(\btheta)$ in the main text.

In addition, there is a relationship between the coordinates $g_i$
of the gradient $\bg = (g_1, \cdots, g_P)$ and the differential 
$\frac{\dd f}{\dd \btheta}(\btheta)$:
\begin{align}
	\forall i \in \{1, \cdots , P\}, \quad 
	g_i = \frac{\dd f}{\dd \btheta}(\btheta)(\mathbf{e}_i) ,
\end{align}
where $\mathbf{e}_i = (0, \cdots, 0, 1, 0, \cdots, 0) \in \mathbb{R}^P$ is the $i$-th
vector of the canonical basis (in other words, $\mathbf{e}_i$ is the one-hot representation
of the integer $i$).

And, of course, the $g_i$ can be calculated by using the partial derivatives:
\begin{align}
	\forall i \in \{1, \cdots , P\}, \quad 
	g_i = \frac{\partial f}{\partial \theta_i}(\btheta) .
\end{align}

\paragraph{Differential of order $d$.}
We suppose that the differential of order $d-1$ of $f$ at $\btheta$ is well-defined
and is a $(d-1)$-linear form on $\mathbb{R}^P$.
We denote it by:
\begin{align}
	\frac{\dd^{d-1} f}{\dd \btheta^{d-1}}(\btheta) \in \mathcal{L}_{d-1}( \mathbb{R}^P,  \mathbb{R}) .
\end{align}

Thus, one can apply $\frac{\dd^{d-1} f}{\dd \btheta^{d-1}}(\btheta)$
to a sequence of vectors $(\bu^1, \cdots, \bu^{d-1})$. We can write:
\begin{align}
	\frac{\dd^{d-1} f}{\dd \btheta^{d-1}}(\btheta) : 
	& \;\mathbb{R}^P \times \cdots \times \mathbb{R}^P \rightarrow \mathbb{R} \nonumber\\
	&(\bu^1, \cdots, \bu^{d-1}) \mapsto \frac{\dd^{d-1} f}{\dd \btheta^{d-1}}(\btheta)(
	\bu^1, \cdots, \bu^{d-1}) .
\end{align}

Now, given a sequence of vectors $(\bu^1, \cdots, \bu^{d-1})$,
let us define $g(\cdot)[\bu^1, \cdots, \bu^{d-1}] : \mathbb{R}^P \rightarrow \mathbb{R}$ such that:
\begin{align}
	g(\btheta)[\bu^1, \cdots, \bu^{d-1}] = \frac{\dd^{d-1} f}{\dd \btheta^{d-1}}(\btheta)(
	\bu^1, \cdots, \bu^{d-1}) .
\end{align}
So, $g(\cdot)[\bu^1, \cdots, \bu^{d-1}]$ is a function from $\mathbb{R}^P$ to
$\mathbb{R}$, and $g(\btheta)[\cdot] \in \mathcal{L}_{d-1}(\mathbb{R}^P, \mathbb{R})$.

As a smooth function from $\mathbb{R}^P$ to $\mathbb{R}$,
one can compute the differential of $g(\cdot)[\bu^1, \cdots, \bu^{d-1}]$ at $\btheta$, that
is a linear form:
\begin{align}
	\frac{\dd g}{\dd \btheta}(\btheta)[\bu^1, \cdots, \bu^{d-1}] :&\; \mathbb{R}^P \rightarrow \mathbb{R} \nonumber \\
	& \bu^d \mapsto h(\btheta)(\bu^d) = \frac{\dd g}{\dd \btheta}(\btheta)[\bu^1, \cdots, \bu^{d-1}](\bu^d).
\end{align}

We change the notation slightly by setting:
\begin{align}
	\frac{\dd g}{\dd \btheta}(\btheta)[\bu^1, \cdots, \bu^{d-1}, \bu^d] :=
	\frac{\dd g}{\dd \btheta}(\btheta)[\bu^1, \cdots, \bu^{d-1}](\bu^d) .
\end{align}
With this notation, it can be proven that 
$\frac{\dd g}{\dd \btheta}(\btheta)[\cdot]$ is a $d$-linear form 
(it belongs to $\mathcal{L}_d(\mathbb{R}^P, \mathbb{R})$).
Finally, by definition of $g$:
\begin{align}
	\frac{\dd g}{\dd \btheta}(\btheta)[\bu^1, \cdots, \bu^{d-1}, \bu^d]
	= \frac{\dd}{\dd \btheta}\left(\frac{\dd^{d-1} f}{\dd \btheta^{d-1}}(\btheta)(
	\bu^1, \cdots, \bu^{d-1})\right)(\bu^d) ,
\end{align}
that we denote by:
\begin{align}
	\frac{\dd^{d} f}{\dd \btheta^{d}}(\btheta)[
	\bu^1, \cdots, \bu^{d-1}, \bu^{d}].
\end{align}
So, $\frac{\dd^{d} f}{\dd \btheta^{d}}(\btheta) \in \mathcal{L}_d(\mathbb{R}^P, \mathbb{R})$.

Like the order-$1$ differential, the order-$d$ differential
can be represented by a tensor.
For instance, a canonical representation of $\frac{\dd^{d} f}{\dd \btheta^{d}}(\btheta)$
is $\bT \in \mathbb{R}^{P^d}$ with:
\begin{align}
	T_{i_1, \cdots, i_d} = 
	\frac{\dd^{d} f}{\dd \btheta^{d}}(\btheta)[\mathbf{e}_{i_1}, \cdots, \mathbf{e}_{i_d}] \in \mathbb{R} ,
\end{align}
where $T_{i_1, \cdots, i_d}$ is the value located at index $(i_1, \cdots, i_d)$ in $\bT$.

We can also define $\bT$ with partial derivatives:
\begin{align}
	T_{i_1, \cdots, i_d} = 
	\frac{\partial^{d} f}{\partial \theta_{i_1} \cdots \partial \theta_{i_d}}(\btheta) \in \mathbb{R} .
\end{align}

\paragraph{Example with $d = 2$.}
With $d = 2$, the tensor $\bT$ representing the order-$2$ differential is the Hessian matrix.
So, $\bT \in \mathbb{R}^{P^2}$ with:
\begin{align}
	T_{ij} = \frac{\dd^{2} f}{\dd \btheta^{2}}(\btheta)[\mathbf{e}_{i}, \mathbf{e}_{j}]
	= \frac{\partial^{2} f}{\partial \theta_{i} \partial \theta_{j}}(\btheta) .
\end{align}

\subsection{Partial derivatives with respect to vectors}

We also need to define formally the following notation,
used in Section \ref{sec:short:higher_order}:
\begin{align}
	\frac{\partial^d f}{\partial \bT^{i_1} \cdots \partial \bT^{i_d}}(\btheta).
\end{align}

Without loss of generality, we only consider the case where
the $\bT^i$ are vectors (and not higher-order tensors).

\paragraph{Representation of $\btheta$ as a sequence of vectors.}
We consider that the argument $\btheta \in \mathbb{R}^P$ of the 
function $f$ can be represented as a sequence of $S$ vectors.
For instance:
\begin{align}
	\btheta = (\theta_1, \cdots, \theta_P) &\cong ((\theta_1, \theta_3, \theta_5, \cdots),
	(\theta_2, \theta_4, \theta_6, \cdots)), \\
	\text{or } \; \btheta = (\theta_1, \cdots, \theta_P) &\cong ((\theta_1, \theta_2, \cdots, \theta_{P_1}),
	(\theta_{P_1 + 1}, \cdots, \theta_{P_1 + P_2}), \nonumber \\
	&\hphantom{=}\;\;(\theta_{P_1 + P_2 + 1}, \cdots, \theta_{P_1 + P_2 + P_3}), \cdots ), \\
	\text{etc., }\;\hphantom{\btheta = (\theta_1, \cdots, \theta_P) }& \nonumber
\end{align}
where $P_1, P_2, \cdots, P_S$ are integers such that $P_1 + \cdots + P_S = P$,
and ``$\cong$'' means ``is represented by''. 
It is essential that each $\theta_i$ appears exactly once in the
right-hand side of the equations above.

Without loss of generality, $\btheta$ can be represented by a sequence of $S$ vectors
with defined sizes $(P_1, \cdots, P_S)$:
\begin{align}
	\btheta \cong (\bT^1, \bT^2, \cdots, \bT^S) \in \mathbb{R}^{P_1} \times 
	\mathbb{R}^{P_2} \times \cdots \times  \mathbb{R}^{P_S} .
\end{align}

\paragraph{Single partial derivative.}

Let $\bu \in \mathbb{R}^P$ be a vector. Just as for $\btheta$,
we represent $\bu$ by a sequence of vectors:
\begin{align}
	\bu \cong (\bU^1, \bU^2, \cdots, \bU^S) \in \mathbb{R}^{P_1} \times 
	\mathbb{R}^{P_2} \times \cdots \times  \mathbb{R}^{P_S} .
\end{align}
To be more specific, if
$\bT^i$ contains $(\theta_1, \theta_3, \theta_6)$,
then $\bU^i$ contains $(u_1, u_3, u_6)$.

Then, we can define $\frac{\partial f}{\partial \bT^{i}}(\btheta)$ as
a linear form belonging to $\mathcal{L}(\mathbb{R}^{P_i}, \mathbb{R})$
with the following property:
\begin{align}
	\frac{\partial f}{\partial \bT^{i}}(\btheta) :\;&
	\mathbb{R}^{P_i} \rightarrow \mathbb{R} \\
	&\bU^i \mapsto \frac{\partial f}{\partial \bT^{i}}(\btheta)[\bU^i]
	= \sum_{k = 1}^{P_i} \frac{\partial f}{\partial T^i_k}(\btheta)
	U^i_k ,
\end{align}
where $T^i_k$ is the $k$-th coordinate of $\bT^i$
and $U^i_k$ is the $k$-th coordinate of $\bU^i$. To be more
specific, if $T^i_k$ represents $\theta_{q}$,
then $\frac{\partial f}{\partial T^i_k}(\btheta)
= \frac{\partial f}{\partial \theta_q}(\btheta)$.

\paragraph{Multiple partial derivatives.}
We can define $\frac{\partial^d f}{\partial \bT^{i_1} \cdots \partial \bT^{i_d}}(\btheta)$ as
a $d$-linear form belonging to 
$\mathcal{L}(\mathbb{R}^{P_{i_1}} \times \cdots \times \mathbb{R}^{P_{i_d}}, \mathbb{R})$
with the following property:
\begin{align}
	\frac{\partial^d f}{\partial \bT^{i_1} \cdots \partial \bT^{i_d}}(\btheta) :\;&
	\mathbb{R}^{P_{i_1}} \times \cdots \times \mathbb{R}^{P_{i_d}} \rightarrow \mathbb{R} \nonumber \\
	&(\bU^{i_1}, \cdots, \bU^{i_d}) \mapsto \frac{\partial^d f}{\partial \bT^{i_1} \cdots \partial \bT^{i_d}}
	(\btheta)[\bU^{i_1}, \cdots, \bU^{i_d}] \nonumber \\
	&\hphantom{(\bU^{i_1}, \cdots, \bU^{i_d}) \mapsto}
	= \sum_{k_1 = 1}^{P_1} \cdots \sum_{k_d = 1}^{P_d} 
	\frac{\partial f}{\partial T^{i_1}_{k_1} \cdots \partial T^{i_d}_{k_d}}(\btheta) 
	U^{i_1}_{k_1}  \cdots  U^{i_d}_{k_d} .
\end{align}

\section{Comparison of training times} \label{app:times}

In Table \ref{tbl:comparison_times}, we report the training times
of 4 different neural networks with Adam, K-FAC and our method. 
Each value is the training time (wall-time) of the configuration in seconds,
averaged over 5 runs.
Note that MLP and LeNet were trained over 200 epochs, 
while BigMLP and VGG were trained over 100 epochs, which 
explains why the training times are larger for smaller networks.

For small networks (MLP, LeNet), the training times are very close
with the different optimizers. However, we observe significant differences 
with large networks (BigMLP, VGG): compared to Adam, the training
is 2 times longer with K-FAC and 3 times longer with our method.

Thus, the computational overhead of our method is either
very small or not excessively large compared to K-FAC.

\begin{table}[h]
	\caption{Comparison of training times (in seconds) of different optimization techniques for
		the 4 main setups.}
	\label{tbl:comparison_times}
	\vspace*{3mm}
	\centering
	\begin{tabular}{lccc}
		\toprule
		Setup & Adam & K-FAC & Ours \\
		\midrule 
		MLP & 2848 & 2953 & 3315 \\
		LeNet & 2944 & 3022 & 3369 \\
		BigMLP & 1777 & 2989 & 4365 \\
		VGG & 1696 & 3117 & 4613 \\
		\bottomrule
	\end{tabular}
\end{table}

\end{document}